\documentclass{article}

\usepackage{PRIMEarxiv}

\usepackage[utf8]{inputenc} 
\usepackage[T1]{fontenc}    
\usepackage{hyperref}       
\usepackage{url}            
\usepackage{booktabs}       
\usepackage{amsfonts}       
\usepackage{nicefrac}       
\usepackage{microtype}      
\usepackage{lipsum}
\usepackage{fancyhdr}       
\usepackage{graphicx}       
\graphicspath{{media/}}     

\usepackage[utf8]{inputenc} 
\usepackage[T1]{fontenc}    
\usepackage{hyperref}       
\usepackage{url}            
\usepackage{booktabs}       
\usepackage{amsfonts}       
\usepackage{nicefrac}       
\usepackage{microtype}      
\usepackage{xcolor}         
\usepackage{subcaption}     
\usepackage{booktabs}       
\usepackage{multirow}       
\usepackage{tabularx}       
\usepackage{pifont}         
\newcommand{\cmark}{\ding{51}} 
\newcommand{\xmark}{\ding{55}} 
\usepackage{rotating}       
\usepackage{makecell}       
\usepackage{wrapfig}
\usepackage{nameref}        
\newcommand{\secref}[1]{\autoref{#1}}

\def\openreview{\url{https://openreview.net/forum?id=jr3nlhNvO4}}

\title{Retail-786k: a Large-Scale Dataset\\ for Visual Entity Matching
}

\author{
  Bianca Lamm\\
  Markant Services International GmbH\\
  Offenburg, Germany\\
  \texttt{Bianca.Lamm@de.markant.com}\\
   \And
  Janis Keuper\\
  Institute for Machine Learning and Analytics (IMLA)\\
  Offenburg University, Germany \\
  \texttt{keuper@imla.ai} \\
}

\begin{document}
\maketitle

\begin{abstract}
Entity Matching (EM) defines the task of learning to group objects by transferring semantic concepts from example groups (=entities) to unseen data. Despite the general availability of image data in the context of many EM-problems, most currently available EM-algorithms solely rely on (textual) meta data.\\
In this paper, we introduce the first publicly available large-scale dataset for ``visual entity matching'', based on a production level use case in the retail domain. Using scanned advertisement leaflets, collected over several years from different European retailers, we provide a total of $\sim$786k manually annotated, high resolution product images containing $\sim$18k different individual retail products which are grouped into $\sim$3k entities. The annotation of these product entities is based on a price comparison task, where each entity forms an equivalence class of comparable products.\\
Following on a first baseline evaluation, we show that the proposed ``visual entity matching'' constitutes a novel learning problem which can not sufficiently be solved using standard image based classification and retrieval algorithms. Instead, novel approaches which allow to transfer example based visual equivalent classes to new data are needed to address the proposed problem. The aim of this paper is to provide a benchmark for such algorithms.\\
\\
Information about the dataset, evaluation code and download instructions are provided under the website: \url{https://www.retail-786k.org/}.
\end{abstract}


\section{Introduction}
\label{sec:introduction}

\noindent {\it Entity Matching} (EM) \cite{barlaug2021neural}, also known as {\it Record Linkage} or {\it Fuzzy Matching} in literature, describes the task of identifying semantically grouped objects across diverse data sources. One of the main applications of EM arises in the context of automated product price monitoring, which aims at the comparability of product prices between different retailers and exchangeable products. For example, a common setting is to match large numbers of products from different online shops against a user defined semantic product entity which is grouping products by certain properties. Most of the available product matching approaches are currently solely based on textual information extracted from product descriptions (e.g., in online shops) and available meta data from product data bases. However, in many practical cases, additional product images could be utilized to improve the semantic matching beyond fixed meta data categories.
\begin{figure}
    \centering
    \def\svgwidth{\columnwidth}
    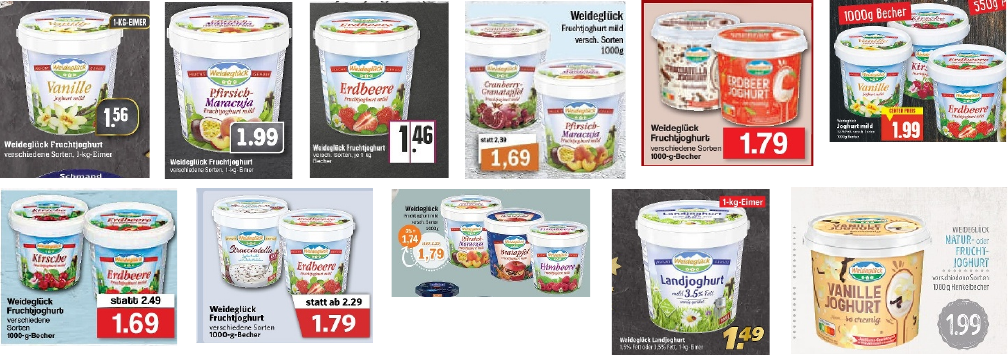
    \caption{In the context of retail products, the term ``visual entity matching'' refers to the task of linking individual product images from diverse sources to a semantic product grouping. Here all images show different products from the same entity which is defined by the fact that single images are used as ``placeholders`` by retailers to promote all products of the entity.  
    For higher resolution version, refer to  Figure\,\ref{fig:appendix_visual_abstract_explain} in the appendix.
    }
    \label{fig:visual_abstract_explain}
\end{figure}
\subsection{Definitions}
\label{subsec:definitions}
Before elaborating into the properties of visual entities, some basic definitions are needed:\\
{\bf Products.} Every retail item on sale is uniquely identified by the internationally standardized {\it Global Trade Item Number} (GTIN).\footnote{https://www.gtin.info/} This is a 14-digit number which differs for any possible variation of an item, e.g., different package designs or packaging levels. We follow the definition that a product is an item with a unique GTIN. Figure\,\ref{fig:product_examples} in the appendix shows three different products with their GTIN. Different products might be very similar or even visually indistinguishable and the application of image classification methods on products often results in {\it fine-grained image classification} (FC) \cite{wei2021fine} problems.\\
{\bf Entities} are then defined as semantic groupings of products, creating an equivalence class over products induced by some external tasks, e.g. price comparison. In the context of the proposed dataset, an entity represents a group of different GTINs,~i.e., contains multiple products which are used by different retailers as equivalent placeholders to promote all of the different products. For example, the three different products in Figure\,\ref{fig:product_examples} belong to the same entity. A subset of the samples of this entity is illustrated in Figure\,\ref{fig:visual_abstract_explain}.\\
\noindent \textbf{Visual Product Entity Matching.} Extending product matching by the extraction and semantic comparison of image features is a straight forward extension of existing approaches, which receives growing attention. From a computer vision perspective, visual EM is closely related to image retrieval \cite{chen2022deep} and fine-grained, long-tail image classification. However, the way that many product entities are defined for applications like price monitoring, demand algorithms to handle different similarity measures and visual intra-entity variances. Figure\,\ref{fig:visual_abstract_explain} illustrates an example for a visual EM instance with a high visual intra-entity variance. In the context of the practical retail task, entities are ”things that should be compared” (mostly in price). The actual task is to learn these entities by examples, forming a novel learning problem.
\subsection{A Dataset for Visual Entity Matching.}
Despite the popularity and practical importance of EM, there are currently only a few publicly available datasets which provide sufficient image data to further drive research and development in this area. Therefore, we propose {\it Retail-786k}, a large open dataset with over 786k manually annotated images of more than 18.8k products in 3,298 different entities. This dataset has been created from large collections of product advertisements in printed leaflets issued by about 130 different European retail chains over a period of two years. It contains high resolution images of typical retail products. The manual entity annotations map the semantic grouping of a commercial price monitoring application, where entities represent groups of products between which prices should be compared. The promotion for all flavors of a yoghurt (strawberry, vanilla, cherry, or raspberry) in one advertisement differs among retailers because different retailers are depicting different flavors as product placeholder in their advertisements. Such different advertisements are illustrated in Figure\,\ref{fig:visual_abstract_explain}.
In contrast to classification, which is identifying predefined discrete subdivisions of classes, the ``visual entity matching'' task includes a strong transfer learning problem: given only abstract samples of entities as sets of ``things that should belong together``, the objective is to learn this metric and transfer it to unseen objects. Hence, the challenges in EM do not come from high variance within entities and image backgrounds, but from a combination of fine-grained classification (very similar entities) and retrieval (unknown matching metric). Both tasks, image classification and image retrieval, do not solve the visual EM problem like our baseline experiments show (see Section\,\ref{sec:baseline_results}).

\noindent \textbf{The Aim of the Dataset:} the data of our dataset represents a highly relevant practical task in retail. Retailers have a significant interest about the product pricing to be competitive. Before comparing prices the promotions of the products have to be predicted to belong to the same group and entity, respectively. For this task our dataset was created. The EM task differs significantly from standard classification problems and thus provides an interesting new CV problem. We provide the first large-scale dataset to address these research questions and with our dataset the automatization of the product price monitoring can be pushed.

\section{Related Work}
\label{sec:related_work}
We focus our literature review on publicly available and annotated datasets that are utilized in the retail context. Here, most image datasets are used to solve in the context of fine-grained classification (FC) task. Regarding entity matching (EM), there are only a few datasets containing images. The images used in the datasets can be roughly grouped into two categories: staged ”studio” images recorded in controlled environments and ”in the wild” images. The Table\,\ref{tab:appendix_relatedWork_01} and the Table\,\ref{tab:appendix_relatedWork_02} in the appendix give a comprehensive overview over all works discussed in the following paragraphs and additional datasets.\\
\noindent\textbf{Online Shopping Products with Images.} The {\it Retail Product Categorisation Dataset} \cite{elayanithottathil2021retail} covers about 48k products in 21 categories. The images of the products have been recorded in controlled environments. 
The {\it iMaterialist} competition at FGVC5 \cite{iMaterialistFGVC2018} created a fine-grained visual categorization dataset of furniture. The competition used staged ”studio” images.\\
\noindent\textbf{Product Images in the Wild.} Most images of the previous described datasets are highly optimized studio productions. For more realistic real-life scenarios, the {\it Grozi-120} dataset provides 676 in-studio product images from websites and 4,973 images taken from videos of retail items recorded inside a grocery store \cite{merler2007recognizing}. It contains 120 grocery products. An extension of this dataset is {\it Grozi-3.2k} \cite{george2014recognizing}, offering more products and higher-quality images. {\it Products-10k} \cite{bai2020products} contains about 10k product classes for about 150k in-shop photos and customer images with human annotations.\\
\noindent\textbf{Shelf Images.} \cite{peng2020rp2k, karlinsky2017fine, goldman2019precise} provide datasets containing images of supermarket shelves. Consequently, the datasets consist of ”in the wild” images. 
The {\it Retail-121} dataset \cite{karlinsky2017fine} consists of 121 fine-grained retail product categories. The {\it RP2K} dataset \cite{peng2020rp2k} provides a combination of shelf images (about 10k) and product images (about 380k).\\
\noindent\textbf{Datasets for Product Entity Matching.} While there are several available large datasets for product entity matching, e.g.,~\cite{kopcke2010evaluation} and \cite{magellandata}, most of them are solely relying on textual or tabular data. Only a few public sources are providing additional image data: like \cite{plummer2015flickr30k} or \cite{wilke2021towards}.
The \textit{Flickr30K Entities} dataset contains textual image descriptions for "in the wild" images that do not have any relation to retail products. The dataset from \cite{wilke2021towards} is based on image URLs provided by the {\it WDC Product Categorization Goldstandard} \cite{WDCProductCategorizationGoldstandard} dataset. The images are grouped into the staged "studio" category. The {\it Stanford Online Products} dataset\cite{oh2016deep} is built of product photos from the online e-commerce website {\it eBay.com}. It contains 120,053 images in total and 22,634 classes.\\
Despite the large number of diverse datasets discussed above, only a small fraction is suitable for the development and evaluation of visual EM-algorithms, as most only provide annotations for (fine-grained) image classification. The few which provide appropriate settings, are mainly based on textual data. Further, the datasets that actually provide images data, offer much smaller numbers of images and classes than our proposed dataset.
\section{Dataset Description}
\label{sec:dataset_dscrptn}
\subsection{Data Sources}
\label{subsec:data_sources}
The starting point of the dataset creation was a large collection of full page leaflet images from about 130 random well-known European retailers between the years 2016 and 2022 provided by the company Markant Services International GmbH. These source images in JPG format either have been obtained by high resolution scans of physically collected leaflets, or have been extracted from publicly available digital leaflets. Large retail chains often consists of different subsidiaries. Such companies potentially use the same product images in their leaflets. We grouped the subsidiaries of a retail chain to avoid the poisoning of the test sets of each entity. 
The mainly promoted products in leaflets are food and beverages. But household goods, cosmetics, pet foods, or (small) electric devices are also available. Figure\,\ref{fig:leaflet_examples} in the appendix gives three representative samples of leaflet pages from different retailers.
\subsection{Data Creation}
\label{subsec:data_creation}
Multiple steps were necessary to create our dataset. The first step was the manual segmentation of the leaflet page into so-called product information boxes. Each box must consist of the product image, the price, and the product description. Some additional information like logos, price tags, quality seals, or other icons can be included inside a box. Such step was done by a team of 23 persons over about two years. This procedure is integrated in a productive process of a price monitoring tool. After the segmentation process, each product information box is cropped from the leaflet page. Such cropped images form the base of our dataset. The last step of the dataset creation is the actual grouping of the images into entities: each entity represents a group of products with a set of according GTINs. The original leaflet pages are not included in our dataset. Also, neither text information like price, brand, nor description of the promotion are included in the dataset. Furthermore, text detection or recognition annotations of the information boxes on the leaflet pages are also not stored.
\subsection{Dataset in Numbers}
\noindent The dataset consists of 786,179 images labeled with 3,298 different entities. The images are split into sub-sets of 748,715 training images and 37,464 test images by the following constraints: I) each entity in the training set must have at least 10 samples, II) each entity in the test set must have at least 3 samples, III) in order to prevent data pollution, images from one retailer can only exist in the training or in the test set for each considered entity, and IV) the minimum resolution of the images is $82$ pixels in width and $129$ pixels in height while the longer edge is always fixed to $512$. This dataset has a size of $65.6$ GiB. We also provide the same dataset with lower resolution (with a long edge size of $256$), which results in a total size of $22.6$ GiB.
Figure\,\ref{fig:histogram_train_test_images_sort_by_train} in the appendix illustrates the resulting unbalanced distribution of the images in training and test set for all entities descending sorted by the training entities. When splitting images into training and test set, we guarantee that no images from the same retailers or subsidiaries exist in both split sets for each entity in order to avoid test set poisoning.
%
\subsection{Image Properties and Variance}
\label{subsec:image_properties_variance}
The images of the products within an entity can be similar and diverse at the same time. We observe high intra-entity variances which differ from typical natural images dominantly used in the computer vision literature: I) Each retailer has its leaflet design. Thus, the background of the images can be very different, varying from a uniform colored bright white, or dark black background over natural background images to more abstract textures. II) Products of an entity differ usually in flavor. The advertisement of such a variance of products also appears in the number and/or the choice of the product placeholders of the promotion image.
III) The imaging perspectives,~i.e., the point of view, of a product image underlie strong variations. If there are multiple products presented in an image, the perspectives of each might differ, too. Also, the alignment of the product can be different. IV) The color distribution of products often varies strongly between different leaflets. Nevertheless, images from different entities can also have strong similarities. There is a typically very high number of entities with very low extra-entity variations,~i.e., many different entities might be visually very similar. 
\begin{wrapfigure}{r}{0.5\linewidth}
    \centering
    \def\svgwidth{0.5\textwidth}
    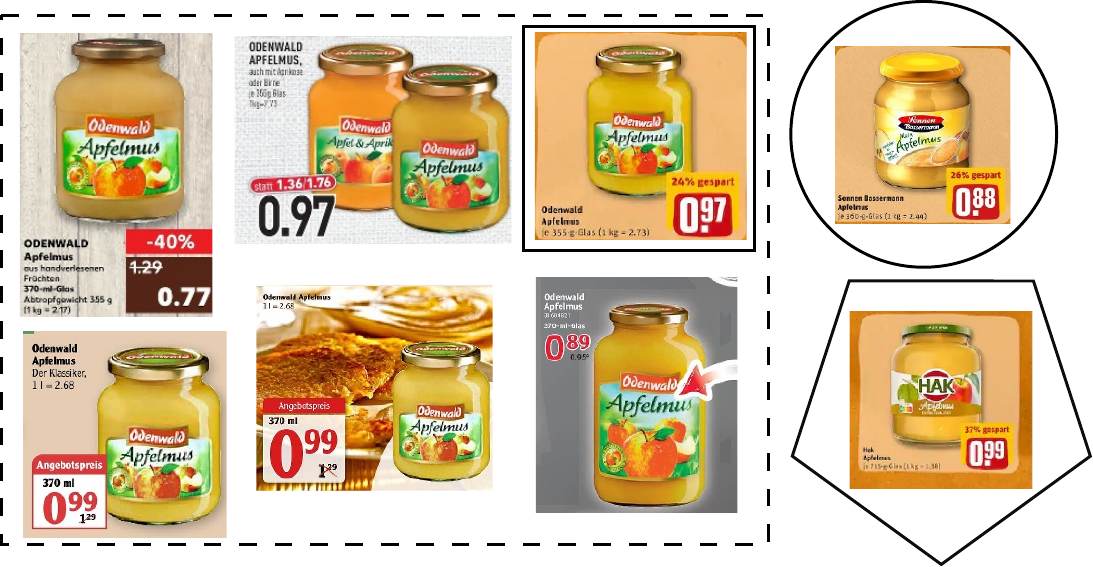
    \caption{Inside the dashed rectangle, samples from a representative entity show the typical intra-entity variations. In contrast to this, there are samples having a strong visually similarity although they belong to different entities. These samples are framed by the solid rectangle, circle, and polygon.}
    \label{fig:product_color_distribution}
\end{wrapfigure}
For instance, the images presented in full resolution in Figure\,\ref{fig:appendx_similarity_Apfelmuss_Odenwald}, Figure\,\ref{fig:appendx_similarity_Apfelmuss_SonnenBassermann}, and Figure\,\ref{fig:appendx_similarity_Apfelmus_HAK} in the appendix show the training sets of three different entities. 
All entities represent the product "apple sauce". The brands of the products as well as the packaging sizes differ for each entity. One special feature for the product "apple sauce" is the different given measure units, g or ml, in the advertisement.
Figure\,\ref{fig:product_color_distribution} shows an example for an intra-entity variance along with two samples of a low extra-entity variance. The images inside the dashed rectangle form an entity. The promotions for this entity vary strongly. However, an image of this entity has a strong similarity to images of other entities. Such an occurrence is illustrated by the images in the solid rectangle, circle, and polygon. Figure\,\ref{fig:entity_example_chips} in the appendix shows a further example of samples from different entities although there is a strong visual similarity. The difference between these entities are the package size of the products.
\section{Baseline Results}
\label{sec:baseline_results}
\noindent We conduct two baseline experiments: First, we treat the proposed EM problem as (fine-grained) classification task. In the second, more realistic dynamic case, we evaluate an image retrieval approach. As discussed, both approaches are theoretically not well suited to actually solve the EM problem and are merely provided to show that standard computer vision algorithms fail to provide the results needed to atomize practically highly relevant EM tasks like price comparison.
\subsection{EM as Classification Problem}
\label{subsec:EM_cls}
We take our dataset as basis for a static classification problem,~i.e., the entities are fixed.
We chose three different state-of-the-art image classification models: {\it ResNet50} \cite{he2016deep}, a {\it Vision Transformer} model \cite{dosovitskiy2020image}, and {\it ConvNeXt} \cite{liu2022convnet}. The specific configuration settings are described in Section\,\ref{subsec:appendix_baseline_EM_classification}. The results are shown in Table\,\ref{tab:baseline_EM_static} in the appendix. Overall, the {\it ConvNeXt} model attains the best test set accuracy of $0.855$.
\subsection{EM as Retrieval Problem} 
\label{subsec:EM_retrieval}
\begin{figure}[!h]
     \centering
     \begin{subfigure}[b]{0.49\textwidth}
         \centering
         \includegraphics[width=\textwidth]{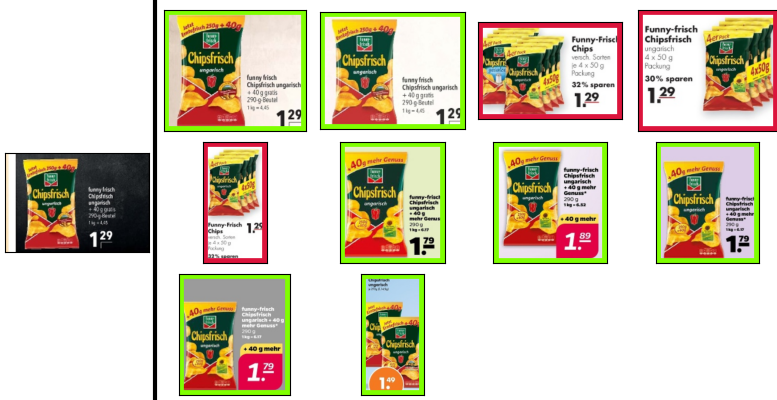}
         \subcaption{Package size}
     \end{subfigure}
     \begin{subfigure}[b]{0.49\textwidth}
         \centering
         \includegraphics[width=\textwidth]{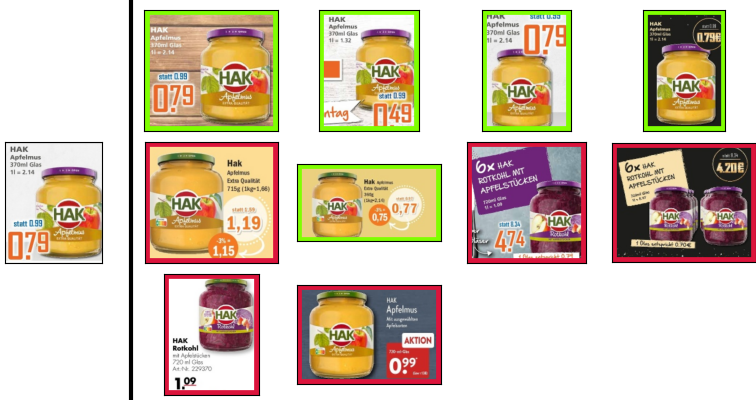}
         \subcaption{Package size and/or product itself}
     \end{subfigure}
     \caption{Illustration of retrieval samples showing the difficulty of matching products. The green-framed images belong to the same entity as the query image. The red-framed images belong to another entity. The sub-captions describe the difference between the query image and the false matched images.
     The images are presented in full resolution in Figure\,\ref{fig:appendix_image_retrieval_Chips} and Figure\,\ref{fig:appendix_image_retrieval_Apfelmus} in the appendix.
     }
     \label{fig:image_retrieval_Chips_Apfelmus}
\end{figure}

For most real applications of visual entity matching, the entities will rapidly change over time as typical retailers are exchanging about 40\% of their products during a single year. Hence, the formulation of visual EM as image retrieval problem is more realistic than as classification task. For our baseline experiment we used the {\it ROADMAP} approach from \cite{ramzi2021robust}, which originally has been evaluated on the {\it Stanford Online Products} (SOP) dataset\cite{oh2016deep}, to obtain the 10 most similar images from the test set to a given test image. 
The settings for our experiments are described in Section\,\ref{subsec:appendix_baseline_EM_retrieval}.
\cite{musgrave2020metric} introduces the metric {\it mAP@R}.
A further explanation of this metric and a comparison of used dataset in \cite{ramzi2021robust} and our dataset is described in Section\,\ref{subsec:appendix_baseline_EM_retrieval}. On the test set of our dataset, we achieved a {\it mAP@R} score of $72.23\%$. Further, the {\it R@10} score is $56.34\%$.
Figure\,\ref{fig:image_retrieval_Chips_Apfelmus} shows the image matching results of two different query images, where the left most image of the sub-figures is the query image. The ten most similar images to the query image are predicted. The green-framed images belong to the same entity as the query image. The red-framed images belong to another entity. The left figure of Figure\,\ref{fig:image_retrieval_Chips_Apfelmus} shows a query image with seven of ten positive image matching. The false matched samples have different package size than the query image. However, all package bags of the products have a yellow border at the top. Also, two of the three false matched images have the same price than the query image. The right figure of Figure\,\ref{fig:image_retrieval_Chips_Apfelmus} shows also negative matched results. Some of these images represent another product but from the same brand. It is more difficult to differ matches when the difference is the package size. Two of the false matched images have a bigger package size than the query image.
More query examples are provided in the appendix, see Figure\,\ref{fig:appendix_image_retrieval_query_01}, Figure\,\ref{fig:appendix_image_retrieval_query_02}, and Figure\,\ref{fig:appendix_image_retrieval_query_03}.
\newpage
\subsection{Error Analysis}
\label{subsec_error_analysis}
In this section, we go deeper into the error analysis of the two approaches, image classification and image retrieval, respectively.\\
\noindent\textbf{Image Classification.} First, we analyze the best classification model {\it ConvNeXt}. This model achieves a precision and recall score of $0.861$ and $0.846$, respectively. As mentioned in Section\,\ref{subsec:EM_cls}, the test accuracy is $0.855$ and the F1-score reaches $0.832$.
The Precision, Recall, and F1-score metrics were calculated by the macro average, i.e. for each label the metric is calculated and the label imbalance of the dataset is not taken into account. 
Section\,\ref{subsec:appendix_error_analysis_classification} describes further investigations regarding misclassifications of images from different entities.\\
\noindent\textbf{Image Retrieval.}
For image retrieval the metric \textit{Recall@K} is used to evaluate the model. The definition of this metric is "For a single query the \textit{Recall@K} is 1 if a positive instance is in the K nearest neighbors, and 0 otherwise. The Recall@K is then averaged on all the queries."\cite{ramzi2021robust}. For our test dataset, we achieve \textit{Recall@1} score of $44.85$\%, \textit{Recall@10} score of $56.34$\%, and {\it mAP@R} score of $72.23\%$. The scores show obviously that the used approach fails for some cases on our dataset. 
Figure\,\ref{fig:error_analysis_image_retrieval} in the appendix shows an example of image retrieval where the most predictions belong to another entity. The top left image beside the vertical line is red framed, i.e. the nearest image to the query image belongs to an entity where single sparkling wine bottles with a content of 0.2l is promoted.
\section{Discussion and Outlook}
\label{sec:discussion_outlook}
With {\it Retail-786k}, we present the first large-scale visual entity matching dataset that has the potential to become a widely accepted benchmark for retail related image classification and matching tasks.
First baseline evaluations show that the EM task that can not sufficiently solved by standard computer vision algorithms like image classification or image retrieval as the best F1-score of the three classification models achieve barely $83.2\%$. The more realistic image retrieval algorithm even accomplishes only a {\it R@10}-score of $56\%$.\\
\noindent\textbf{Limitations.} Despite our best efforts to create and release a dataset which holds up to the highest standards, there are some known limitations:\\
I) Manual miss-annotations: during the dataset creation process, we eliminated many images which have been labeled wrongly. However, there is no guarantee that we did not miss more and less obvious of these cases.\\ 
II) Missing textual data: many existing EM algorithms use text information, like price, price reduction, and product descriptions which is implicitly contained in our images as well. To this end, we have not yet extracted this information.\\
III) Entity labeling: the entities in our dataset have been defined based on the very specific application of price monitoring in printed advertisements. Other EM tasks might define different groupings of the same products. Hence, it is not clear how good results on our dataset will transfer to other problems.\\
\noindent\textbf{Outlook.} Even though the presented version of our dataset provides a good basis for the development of new approaches towards visual product entity matching, the underlying data has even more potential for further improvements: the extraction of textual information and price tags could further enhance the database toward multi-modal solutions.

\impact{The intentions of the proposed dataset includes to provide a benchmark for novel (visual) EM-algorithms. As in any automation task, the development of strong EM-solutions could have negative impact on the employment of people who are currently performing this task manually. On the positive side, i.e. access to powerful price comparison tools will allow customers (especially those with low income) to purchase groceries at the best price.}

\section*{Reproducibility Statement}
\label{sec:reproducibilty_statement}
Information about the dataset and the downloading steps are provided under the website: \url{https://www.retail-786k.org/}. The code necessary to replicate the experiments discussed in this paper has been made publicly available on the GitHub website: \url{https://github.com/retail786k/retail786k-dataset}. Also, a comprehensive description of the dataset is described in Section\,\secref{sec:appendix_datasheets_for_datasets} Datasheets for Datasets.

\newpage
\bibliographystyle{unsrt}  
\bibliography{references}  

\newpage
\appendix

\section{Dataset Information}

\textbf{Dataset Description.} An comprehensive explanation of our dataset is described in "Datasheets for Datasets", Section\,\ref{sec:appendix_datasheets_for_datasets}.\\
\\
\textbf{License.} Our dataset is licensed under the \textit{Creative Commons Attribution-NonCommercial-NoDerivatives 4.0 International} license.\\
\\
\textbf{Hosting.} The dataset is uploaded to the data repository \url{https://zenodo.org/record/7970567} and the DOI of the dataset is 10.5281/zenodo.7970566.\\
\\
\textbf{Support and Maintenance.} The supporting, hosting and maintenance plan is described in Section\,\ref{subsec:maintenance}.\\
\\
\textbf{Download and Reproducibility.} Information about the dataset and code as well as the downloading steps are provided under the website: \url{https://www.retail-786k.org/}.\\
\\
\textbf{Author responsibility.} As authors, we bear all responsibility in case of any violation of rights during the collection of the data or other work, and will take appropriate action when needed, e.g., to remove data with such issues.

\newpage
\section{Datasheets for Datasets}
\label{sec:appendix_datasheets_for_datasets}


\subsection{Motivation}

\textbf{For what purpose was the dataset created?} Was there a specific task in mind? Was there a specific gap that needed to be filled? Please provide a description.\\
{\fontfamily{cmr}\selectfont
The dataset was created as benchmark for "visual entity matching" tasks. Most existing Entity Matching (EM) approaches solely rely on meta data even if the data basis are products.
}
\\
\\
\textbf{Who created the dataset (e.g., which team, research group) and on behalf of which entity (e.g., company, institution, organization)?}\\
{\fontfamily{cmr}\selectfont
The company Markant Services International GmbH and the Institute of Machine Learning and Analytics from the Offenburg University are involved in the dataset creation. From the company a team of 23 persons worked for about two years on the annotation of the dataset. Also, Janis Keuper and Bianca Lamm from the institute participated on the dataset creation.
}
\\
\\
\textbf{Who funded the creation of the dataset?} If there is an associated grant, please provide the name of the grantor and the grant name and number.\\
{\fontfamily{cmr}\selectfont
The creation costs of about 1 million USD of the dataset was funded by the German company Markant Services International GmbH that provides a price monitoring service.
}
\\
\\
\textbf{Any other comments?}\\
{\fontfamily{cmr}\selectfont
None.
}

\subsection{Composition}

\begin{figure}[!h]
    \centering
    \begin{subfigure}[b]{0.2\textwidth}
        \centering
        \includegraphics[width=\textwidth]{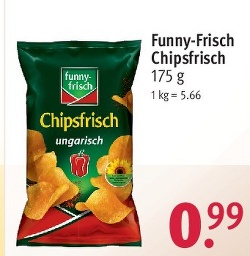}
    \end{subfigure}
    \hspace{10mm}
    \begin{subfigure}[b]{0.2\textwidth}
        \centering
        \includegraphics[width=\textwidth]{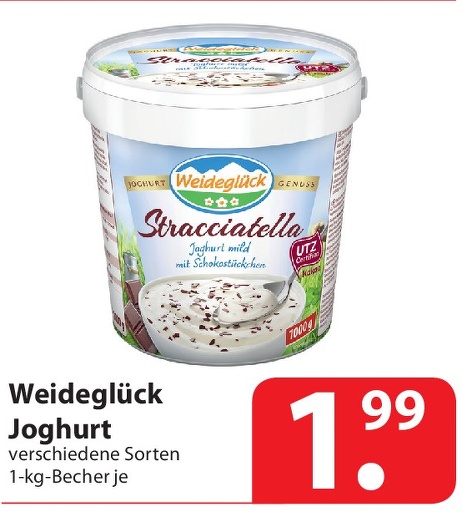}
    \end{subfigure}
    \caption{Illustration of two instances from different groupings.
    }
    \label{fig:appendix_datasheets_instance_examples}
\end{figure}

\textbf{What do the instances that comprise the dataset represent (e.g., documents, photos, people, countries)?} Are there multiple types of instances (e.g., movies, users, and ratings; people and interactions between them; nodes and edges)? Please provide a description.\\
{\fontfamily{cmr}\selectfont
The instances are cropped images in JPG format from German leaflets. Each image promotes one or multiple retail product(s). The instances are grouped in so-called entities being a semantic grouping of products. The entities are sequentially numbered with an ID. Two instances from different groupings are shown in Figure\,\ref{fig:appendix_datasheets_instance_examples}.
}
\\
\textbf{How many instances are there in total (of each type, if appropriate)?}\\
{\fontfamily{cmr}\selectfont
There are 786,179 images in total that are distributed over 3,298 entities (semantic groupings). These images are split into a training set of 748,715 images and a test set of 37,464 images.
}
\\
\\
\textbf{Does the dataset contain all possible instances or is it a sample (not necessarily random) of instances from a larger set?} If the dataset is a sample, then what is the larger set? Is the sample representative of the larger set (e.g., geographic coverage)? If so, please describe how this representativeness was validated/verified. If it is not representative of the larger set, please describe why not (e.g., to cover a more diverse range of instances, because instances were withheld or unavailable).\\
{\fontfamily{cmr}\selectfont
The dataset contains all possible instances because no more products were considered. Therefore, there are no more cropped images from leaflets of other products.
}
\\
\\
\newpage
\textbf{What data does each instance consist of?} “Raw” data (e.g., unprocessed text or images) or features? In either case, please provide a description.\\
{\fontfamily{cmr}\selectfont
An instance is a cropped image from a leaflet. These leaflets either have been obtained by high resolution scans of physically collected leaflets, or have been extracted from publicly available digital leaflets. At the resulted images in JPG format are unprocessed images.
}
\\
\\
\textbf{Is there a label or target associated with each instance?} If so, please provide a description.\\
{\fontfamily{cmr}\selectfont
The label is the entity ID that is given by the folder name in which the instance is stored.
}
\\
\\
\textbf{Is any information missing from individual instances?} If so, please provide a description, explaining why this information is missing (e.g., because it was unavailable). This does not include intentionally removed information, but might include, e.g., redacted text.\\
{\fontfamily{cmr}\selectfont
There is no text extraction of the product descriptions from each promotion image. The wording of this text is unavailable. Information about price, discount, brand, and some attributes are not available at the time of the dataset creation. But these information can be received and another version of the dataset could be published.
}
\\
\\
\textbf{Are relationships between individual instances made explicit (e.g., users’ movie ratings, social network links)?} If so, please describe how these relationships are made explicit.\\
{\fontfamily{cmr}\selectfont
An instance, a promotion image, is assigned a list of GTIN from the promoted products. Hence, an entity has a set of GTIN lists. Such GTIN lists are disjunctive for all entities. Inside an entity, an instance must have at least one equal GTIN of its GTIN-list with another instance of this entity.
}
\\
\\
\textbf{Are there recommended data splits (e.g., training, development/validation, testing)?} If so, please provide a description of these splits, explaining the rationale behind them.\\
{\fontfamily{cmr}\selectfont
The data splits are given by the dataset itself. There are training and test splits. No images from the same retailers or subsidiaries exist in both split sets for each entity in order to avoid test set poisoning due to the visual similarity of promotions.
}
\\
\\
\textbf{Are there any errors, sources of noise, or redundancies in the
dataset?} If so, please provide a description.\\
{\fontfamily{cmr}\selectfont
Redundancies can appear in the dataset, e.g. there could exist different entities that represent the same product. This could be caused by false manual classification of GTINs for the promoted products in the instance. 
}
\\
\\
\textbf{Is the dataset self-contained, or does it link to or otherwise rely on external resources (e.g., websites, tweets, other datasets)?} If it links to or relies on external resources, a) are there guarantees that they will exist, and remain constant, over time; b) are there official archival versions of the complete dataset (i.e., including the external resources as they existed at the time the dataset was created); c) are there any restrictions (e.g., licenses, fees) associated with any of the external resources that might apply to a dataset consumer? Please provide descriptions of all external resources and any restrictions associated with them, as well as links or other access points, as appropriate.\\
{\fontfamily{cmr}\selectfont
The dataset is self-contained and does not link to external resources.
}
\\
\\
\textbf{Does the dataset contain data that might be considered confidential (e.g., data that is protected by legal privilege or by doctor–patient confidentiality, data that includes the content of individuals’ non-public communications)?} If so, please provide a description.\\
{\fontfamily{cmr}\selectfont
No, there are no personal data included in the dataset.
}
\\
\\
\textbf{Does the dataset contain data that, if viewed directly, might be offensive, insulting, threatening, or might otherwise cause anxiety?}
If so, please describe why.\\
{\fontfamily{cmr}\selectfont
No, such data does not exist in the dataset.
}
\\
\\
\textbf{Any other comments?}\\
{\fontfamily{cmr}\selectfont
None.
}
\newpage
\subsection{Collection Process}

\textbf{How was the data associated with each instance acquired?} 
Was the data directly observable (e.g., raw text, movie ratings), reported by subjects (e.g., survey responses), or indirectly inferred/derived from other data (e.g., part-of-speech tags, model-based guesses for age or language)? If the data was reported by subjects or indirectly inferred/derived from other data, was the data validated/verified? If so, please describe how.\\
{\fontfamily{cmr}\selectfont
The instances are derived from online available leaflets. The leaflet images were neither validated nor verified.
}
\\
\\
\textbf{What mechanisms or procedures were used to collect the data
(e.g., hardware apparatuses or sensors, manual human curation,
software programs, software APIs)?} How were these mechanisms or
procedures validated?\\
{\fontfamily{cmr}\selectfont
The following procedures were used to collect the data:
\begin{enumerate}
    \item Downloading the public available leaflets of each retailer at each calendar week.
    \item Uploading the digital leaflets into an internal software of the company Markant Services International GmbH.
    The software converts the leaflets from PDF format into JPG format.
    \item The employees of the company draw rectangles onto each leaflet page to manual segment the page into so-called product information boxes. Each box must consist of the product image, the price, and the product description. Some additional information like logos, price tags, quality seals, or other icons can be included inside a box.
    \item The leaflet can be cropped into the instances of the dataset with the stored rectangle coordinates from the previous step.
\end{enumerate}
There was no validation of the procedures, especially for the drawing of the product information boxes.
}
\\
\\
\textbf{If the dataset is a sample from a larger set, what was the sampling strategy (e.g., deterministic, probabilistic with specific sampling probabilities)?}\\
{\fontfamily{cmr}\selectfont
The dataset is not a sample from a larger set.
}
\\
\\
\textbf{Who was involved in the data collection process (e.g., students,
crowdworkers, contractors) and howwere they compensated (e.g.,
how much were crowdworkers paid)?}\\
{\fontfamily{cmr}\selectfont
The company Markant Services International GmbH, who funded the dataset, involve an own team of its employees for the data collection process.
The specific payment cannot be disclosed because of the confidentiality of company internal data. But the is a lower bound given by the legal minimum wage of 12.60 USD per hour in Germany (in 2022). The total amount of costs aggregate about 1m USD.
}
\\
\\
\textbf{Over what timeframe was the data collected?} Does this timeframe
match the creation timeframe of the data associated with the instances
(e.g., recent crawl of old news articles)? If not, please describe the timeframe in which the data associated with the instances was created.\\
{\fontfamily{cmr}\selectfont
The team of 23 persons worked in a period of two years to create the dataset. The basis of the dataset consists of about 68k leaflets. The release years of the considered leaflets and hence the data collection timeframe are from 2016 to 2022.
}
\\
\\
\textbf{Were any ethical review processes conducted (e.g., by an institutional review board)?} If so, please provide a description of these review processes, including the outcomes, as well as a link or other access point to any supporting documentation.\\
{\fontfamily{cmr}\selectfont
Unknown to the authors of the datasheet. But there are no participant risks. There is no personal data collection from the people who are created the dataset.
}
\\
\\
\textbf{Any other comments?}\\
{\fontfamily{cmr}\selectfont
None.
}
\newpage
\subsection{Preprocessing/cleaning/labeling}

\textbf{Was any preprocessing/cleaning/labeling of the data done (e.g., discretization or bucketing, tokenization, part-of-speech tagging, SIFT feature extraction, removal of instances, processing of missing values)?} If so, please provide a description. If not, you may skip the
remaining questions in this section.\\
{\fontfamily{cmr}\selectfont
There are some entities that group arbitrary products due to manual miss-annotations of the GTINs to these products.
Because of this occurrence, cake baking mix, cream, and pudding form an entity.
We eliminated many images that have been labeled wrongly to clean the dataset.
However, there is no guarantee that we did not miss more and less obvious of these cases.
}
\\
\\
\textbf{Was the “raw” data saved in addition to the preprocessed/cleaned/labeled data (e.g., to support unanticipated future uses)?} If so, please provide a link or other access point to the “raw” data.\\
{\fontfamily{cmr}\selectfont
No, the "raw" data is not saved separately. 
The leaflets being the basis for the dataset creation are stored at the company Markant Services International GmbH internally.
}
\\
\\
\textbf{Is the software that was used to preprocess/clean/label the data available?} If so, please provide a link or other access point.\\
{\fontfamily{cmr}\selectfont
No, the cleaning of the data was done by manual screening. There is no code / software available.
}
\\
\\
\textbf{Any other comments?}\\
{\fontfamily{cmr}\selectfont
None.
}

\subsection{Uses}

\textbf{Has the dataset been used for any tasks already?} If so, please provide a description.\\
{\fontfamily{cmr}\selectfont
The dataset itself has not been used for any tasks already. 
But a subset of the dataset was used company internally for the price monitoring service.
}
\\
\\
\textbf{Is there a repository that links to any or all papers or systems that use the dataset?} If so, please provide a link or other access point.\\
{\fontfamily{cmr}\selectfont
No, there is no repository.
}
\\
\\
\textbf{What (other) tasks could the dataset be used for?}\\
{\fontfamily{cmr}\selectfont
The dataset can be also used for the image classification or the image retrieval task. Both tasks are investigated in our experiments. Unfortunately, they do not solve the visual EM problem. Furthermore, the images in the dataset can be used for the OCR task. But the Ground Truth data of the text in the images is not provided by our dataset.
}
\\
\\
\textbf{Is there anything about the composition of the dataset or the way it was collected and preprocessed/cleaned/labeled that might impact future uses?} For example, is there anything that a dataset consumer might need to know to avoid uses that could result in unfair treatment of individuals or groups (e.g., stereotyping, quality of service issues) or other risks or harms (e.g., legal risks, financial harms)? If so, please provide a description. Is there anything a dataset consumer could do to mitigate these risks or harms?\\
{\fontfamily{cmr}\selectfont
Unknown to the authors of the datasheet.
}
\\
\\
\textbf{Are there tasks for which the dataset should not be used?} If so, please provide a description.\\
{\fontfamily{cmr}\selectfont
Unknown to the authors of the datasheet.
But the license of the dataset prohibits commercial purpose and the distribution of a modified version of the dataset.}
\\
\\
\textbf{Any other comments?}\\
{\fontfamily{cmr}\selectfont
None.
}

\subsection{Distribution}

\textbf{Will the dataset be distributed to third parties outside of the entity (e.g., company, institution, organization) on behalf of which the dataset was created?} If so, please provide a description.\\
{\fontfamily{cmr}\selectfont
Yes, the dataset is publicly available on the internet.
}
\\
\\
\newpage
\textbf{How will the dataset will be distributed (e.g., tarball on website, API, GitHub)?} Does the dataset have a digital object identifier (DOI)?\\
{\fontfamily{cmr}\selectfont
Information about the dataset and code as well as the downloading steps are provided under the website: \url{https://www.retail-786k.org/}.
The dataset can be also downloaded directly via the URL \url{https://zenodo.org/record/797056}. 
The DOI of the dataset is 10.5281/zenodo.7970566.
}
\\
\\
\textbf{When will the dataset be distributed?}\\
{\fontfamily{cmr}\selectfont
The dataset was released at Jun 06, 2023.
}
\\
\\
\textbf{Will the dataset be distributed under a copyright or other intellectual
property (IP) license, and/or under applicable terms of use (ToU)?} If so, please describe this license and/or ToU, and provide a link or other access point to, or otherwise reproduce, any relevant licensing terms or ToU, as well as any fees associated with these restrictions.\\
{\fontfamily{cmr}\selectfont
This dataset is licensed under the \textit{Creative Commons Attribution-NonCommercial-NoDerivatives 4.0 International} license. For more information about the license see \url{https://creativecommons.org/licenses/by-nc-nd/4.0/}.
}
\\
\\
\textbf{Have any third parties imposed IP-based or other restrictions on the data associated with the instances?} If so, please describe these restrictions, and provide a link or other access point to, or otherwise reproduce, any relevant licensing terms, as well as any fees associated with these restrictions.\\
{\fontfamily{cmr}\selectfont
Unknown to the authors of the datasheet.
}
\\
\\
\textbf{Do any export controls or other regulatory restrictions apply to the dataset or to individual instances?} If so, please describe these restrictions, and provide a link or other access point to, or otherwise reproduce, any supporting documentation.\\
{\fontfamily{cmr}\selectfont
Unknown to the authors of the datasheet.
}
\\
\\
\textbf{Any other comments?}\\
{\fontfamily{cmr}\selectfont
None.
}

\subsection{Maintenance}
\label{subsec:maintenance}

\textbf{Who will be supporting/hosting/maintaining the dataset?}\\
{\fontfamily{cmr}\selectfont
Bianca Lamm and Janis Keuper from the Institute of Machine Learning and Analytics from the Offenburg University are supporting/maintaining the dataset.
}
\\
\\
\textbf{How can the owner/curator/manager of the dataset be contacted (e.g., email address)?}\\
{\fontfamily{cmr}\selectfont
The curators of the dataset, Bianca Lamm and Janis Keuper, can be contacted via info@Retail-786k.org, bianca.lamm@hs-offenburg.de, and keuper@imla.ai.
}
\\
\\
\textbf{Is there an erratum?} If so, please provide a link or other access point.\\
{\fontfamily{cmr}\selectfont
No.
}
\\
\\
\textbf{Will the dataset be updated (e.g., to correct labeling errors, add new instances, delete instances)?} If so, please describe how often, by whom, and how updates will be communicated to dataset consumers (e.g., mailing list, GitHub)?\\
{\fontfamily{cmr}\selectfont
In the future, another version of the dataset could be published with information about price, discount, brand, and so on of the products per instance.\\
There is no fixed date for the upload of the new version, yet. Also, there will be no fixed periods for updating the dataset.\\
The new versions will be updated by the authors of the datasheet.
The updates to the dataset will be communicated via the dataset website: \url{https://www.retail-786k.org/}.
}
\\
\\
\textbf{If the dataset relates to people, are there applicable limits on the retention of the data associated with the instances (e.g., were the individuals in question told that their data would be retained for a fixed period of time and then deleted)?} If so, please describe these limits and explain how they will be enforced.\\
{\fontfamily{cmr}\selectfont
The dataset does not relate to people.
}
\\
\\
\newpage
\textbf{Will older versions of the dataset continue to be supported/hosted/maintained?} If so, please describe how. If not, please describe how its obsolescence will be communicated to dataset consumers.\\
{\fontfamily{cmr}\selectfont
There are no older versions of the dataset, yet.
If newer versions are available, then old version will still be available.
}
\\
\\
\textbf{If others want to extend/augment/build on/contribute to the dataset, is there a mechanism for them to do so?} If so, please provide a description. Will these contributions be validated/verified? If so, please describe how. If not, why not? Is there a process for communicating/distributing these contributions to dataset consumers? If so, please provide a description.\\
{\fontfamily{cmr}\selectfont
The license of our dataset does not allow to extend/augment/build on/contribute to the dataset. Interested persons can contact the curators via info@Retail-786k.org.
}
\\
\\
\textbf{Any other comments?}\\
{\fontfamily{cmr}\selectfont
None.
}


\newpage
\section{Additional Related Work}
\label{sec:appendix_add_realted_work}
\noindent In addition to the datasets described in Section\,\ref{sec:related_work}, there are further datasets of interest that are described in this section. The Table\,\ref{tab:appendix_relatedWork_01} and the Table\,\ref{tab:appendix_relatedWork_02} summarize all described datasets from this section and Section\,\ref{sec:related_work} in relation to the number of images and number of classes per dataset. Besides, we indicate whether the dataset is suitable for the FC or EM tasks. The last two columns define the image category: staged ”studio” images recorded in controlled environments and ”in the wild” images.\\
\\
\noindent\textbf{Online Shopping Products with Images.} The assignment of attribute labels for product images was the task of the {\it iMaterialist} competition at FGVC4 \cite{iMaterialistFGVC2017}. The \textit{Amazon Review Data} (2018) \cite{ni2019justifying} describes a dataset of products offered by {\it Amazon}. It contains about 233.1m reviews, e.g., ratings, text, helpfulness votes, or user provided product images. Additionally, the dataset consists of metadata for 15.5m products, including product images. The data is comprised in 29 categories, like \textit{Electronics, Books, All Beauty, Home and Kitchen}, or \textit{Pet Supplies}. The {\it Atlas} dataset \cite{umaashankar2019atlas} focuses on clothing images, which were collected from popular Indian e-commerce stores. Image descriptions and pre-defined product taxonomies are also provided in this 52 category dataset.\newline
\newline
\noindent\textbf{Product Images in the Wild.} The {\it AliProducts Challenge} dataset \cite{Le_2020_ECCV} consists of real-life (”in the wild”) image scenarios, e.g.,~products in customer hand, on a table, in a shopping basket, or on a supermarket shelf. The dataset comprises about 50k different products and about 3m images in total. A dataset considering images recorded from hand-held or wearable cameras and showing hand-held grocery products is the {\it SHORT} dataset \cite{rivera2014small}. The training images are of high quality but form only a small set. Oppositely, the test set comprises nearly 135k smartphone-captured images of 30 products.\\
\newline
\noindent\textbf{Shelf Images.} 25 grocery classes with about 5k images are included in the {\it Freiburg Groceries Dataset} \cite{jund2016freiburg}. The dataset from \cite{varol2015toward} comprises more than 5k shelf images consisting of 10 tobacco brands. \cite{karlinsky2017fine} provides the {\it GameStop} dataset that contains of video game store images. The dataset covers about 3.7k video game categories. More than 1m individual product images retrieved from 11,762 store shelf photographs are provided by the {\it SKU-110} dataset \cite{goldman2019precise}. This dataset is intended for object detection in densely packed scenes. The {\it Locount} dataset \cite{Cai2020Locount}, which consists of about 50k images with more than 1.9m object instances in 140 product categories, aims towards localize groups of products with the number of instances.\newline
\newline
\noindent\textbf{Product Checkout Images.} The {\it RPC} dataset \cite{wei2022rpc} includes checkout data. It consists of 200 categories and 83,739 images in total. The images are divided into single-product images for the training set and 30k multi-product checkout images for the test set. The {\it ARC-100} dataset \cite{bukhari2021arc} contains 31k RGB images with a size of $640\times480$ pixel. The images show 100 commonly found retail items in Lahore, Pakistan. The products were lied on a black, matte finish conveyor belt in various logical orientations. The dataset is made in a controlled environment.\newline
\newline
\noindent\textbf{Products in Vending Machines.} Another way to record images of retail products is in vending machines. For the {\it TGFS} dataset \cite{hao2019take} 38k images from self-service vending machines with a resolution of $480\times640$ pixel are collected. The images show the product removal. The dataset comprises 24 fine-grained and 3 coarse classes. \cite{fuchs2019towards} creates a dataset of 300 images illustrating the view on vending machines. There are 15k labeled instances of 90 products in the images. A similar build dataset as before is the {\it HoloSelecta} dataset \cite{fuchs2020holoselecta}. It also contains shelf images of vending machines of about 295 images with 10,035 labeled instances of 109 products. The considered vending machines include food as well as beverage products. The labeled images provide bounding box coordinates and also information about the brand, product name, and the GTIN of the product.\\ 
\newline
\noindent\textbf{Data Bases with Food Images.} More datasets deal with retail products in the food domain. One dataset limits oneself to fruits and vegetables \cite{rocha2010automatic}. This dataset consists of 2,633 images and 15 categories. Moreover, the Food-5K dataset \cite{singla2016food} contains 2.5k food and 2.5k non-food images. The food images illustrate cooked meals on plates and the non-food images show, e.g., humans, animals, flowers, or landscapes. \cite{singla2016food} provides another food image dataset called {\it Food-11}. This dataset has about 16k food images that are grouped in 11 categories.\\
\\
\noindent\textbf{Miscellaneous.} The {\it MVTec D2S} dataset \cite{follmann2018mvtec} provides semantic segmentation annotated images of retail products. The images are recorded in a laboratory environment. Another retail product dataset is {\it SOIL-47} \cite{burianekSOIL47}. The characteristic feature of the product images is the 20 different captured horizontal views. The dataset consists of 47 product categories and each category covers 21 images.

\newpage
\section{Detailed Elaboration}
\label{sec:appendix_detailed_elaboration}

\begin{figure}
    \centering
    \includegraphics[width=0.75\linewidth]{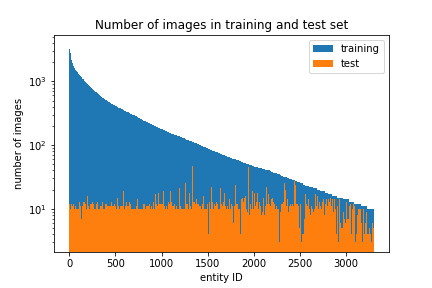}
    \caption{Entity histograms by train and test split of the dataset.}
    \label{fig:histogram_train_test_images_sort_by_train}
\end{figure}

\begin{figure}
    \centering
    \begin{subfigure}[b]{0.2\textwidth}
        \centering
        \includegraphics[width=\textwidth]{images/entity_example/chips_175g_112079.jpg}
        \caption{175g}
    \end{subfigure}
    \hfill
    \begin{subfigure}[b]{0.2\textwidth}
        \centering
        \includegraphics[width=\textwidth]{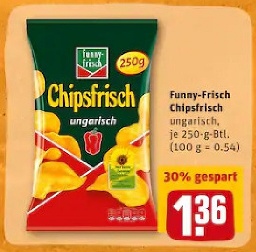}
        \caption{250g}
    \end{subfigure}
    \hfill
    \begin{subfigure}[b]{0.2\textwidth}
        \centering
        \includegraphics[width=\textwidth]{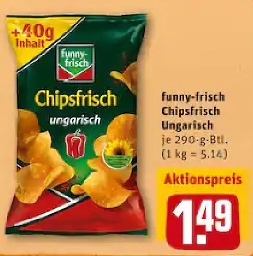}
        \caption{290g}
    \end{subfigure}
    \hfill
    \begin{subfigure}[b]{0.2\textwidth}
        \centering
        \includegraphics[width=\textwidth]{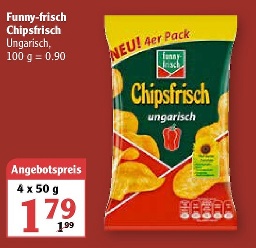}
        \caption{4$\times$50g}
    \end{subfigure}
    \caption{Illustration of samples from different entities that have a strong visual similarity. The difference of the entities is the package size that is specified by the caption of each subfigure.
    For higher resolution versions, refer to  Figure\,\ref{fig:appendix_chips_175g}, \ref{fig:appendix_chips_250g}, \ref{fig:appendix_chips_290g}, and \ref{fig:appendix_chips_4x50g}.
    }
    \label{fig:entity_example_chips}
\end{figure}

Figure\,\ref{fig:histogram_train_test_images_sort_by_train} illustrates the resulting unbalanced distribution of the images in training and test set for all entities descending sorted by the training entities.
Figure\,\ref{fig:entity_example_chips} shows a further example of samples from different entities although there is a strong visual similarity. The difference between these entities are the package size of the products.

\subsection{Baseline Results on EM as Classification Problem}
\label{subsec:appendix_baseline_EM_classification}
We chose three different state-of-the-art image classification models: {\it ResNet50} \cite{he2016deep}, a {\it Vision Transformer} model \cite{dosovitskiy2020image} with a {\it vit$\_$b$\_$16} architecture and {\it ConvNeXt} \cite{liu2022convnet} using the base model architecture. Each model has been pretrained on {\it ImageNet} \cite{deng2009imagenet} and was trained for 50 epochs of fine-tuning. For the {\it ResNet50} model, we used a NVIDIA GeForce RTX 3090 GPU and a batch size of 32. The same batch size is used for the training for the {\it Vision Transformer} model and the {\it ConvNeXt} model. These models were trained on three NVIDIA A100-SXM4-40GB. All models use a SGD optimizer with a learning rate of 0.001 plus a momentum of 0.9. Furthermore, we add a decay of the learning rate of 0.1 every 7 epochs.
Results are shown in Table\,\ref{tab:baseline_EM_static}. The {\it ResNet50} model results in the lowest test set accuracy of $0.824$. An improvement is achieved by the {\it Vision Transformer} model with a test set accuracy of $0.834$. The {\it ConvNeXt} model attains the best test set accuracy of $0.855$. The F1-score values of the test data are also calculated for each model.
\begin{table}[htb]
    \centering
    \caption{Illustration of the results for the ``Static'' EM problem. The models {\it ResNet50, \it Vision Transformer} and {\it ConvNeXt} were examined. The accuracy and the F1-score of the test data as well as the training time in hours are given. (*) Macro average is used.}
    \label{tab:baseline_EM_static}
    \begin{tabular}{cccc}
        \\
        \toprule
         model                      &F1-score* & test accuracy     & training time [h]\\
        \midrule
         {\it ConvNeXt}             &$0.832$        &$0.855$           &$76$\\
         {\it Vision Transformer}   &$0.806$        &$0.834$           &$62$\\
         {\it ResNet50}             &$0.793$        &$0.824$           &$51$\\
         \bottomrule
    \end{tabular}
\end{table}
\subsection{Baseline Results on EM as Retrieval Problem}
\label{subsec:appendix_baseline_EM_retrieval}
We trained on 4 NVIDIA TITAN GPUs for 43 hours, using the following settings for the training: ResNet50 as backbone model, 100 epochs, batch size of 128, Adam optimizer, and an embedding dimension of 512. The settings for the {\it ROADMAP} approach are the same as in the experiments of the paper \cite{ramzi2021robust}. \cite{musgrave2020metric} introduces the metric {\it mAP@R}, where the authors define $R$ as “For each query[...], let $R$ be the total number of references that are the same class as the query” \cite{musgrave2020metric}. Hence, the value of $R$ is different for each query and is therefore not a configurable setting value. The total number of references for each entity is consequently the sum of images of this entity. 
The authors of \cite{ramzi2021robust} investigate the three dataset SOP \cite{oh2016deep}, CUB \cite{wah2011caltech}, and INaturalist \cite{van2018inaturalist}. The last called dataset contains images of animals and comprises 8,142 classes. A specialization to the animal bird is the CUB dataset. It contains 200 fine-grained classes of birds. The SOP dataset has the highest number of classes, 22,634 classes. A comparison of the three datasets and our dataset in particular to the mAP@R score and the number of classes is shown in the Table\,\ref{tab:mAP_at_R_values}. On the test set of our dataset, we achieved a {\it mAP@R} score of $72.23\%$. Further, the {\it R@10} score is $56.34\%$.
\begin{table}[htb]
    \centering
    \caption{Comparison of the number of classes and the {\it mAP@R} value between three image classification datasets and our dataset.
    }
    \label{tab:mAP_at_R_values}
    \begin{tabular}{ccc}
        \\
        \toprule
         dataset        &number of classes  & mAP@R [\%]\\
        \midrule
         CUB            &200                &25.27±0.12\\
         INaturalist    &8,142              &27.01±0.10\\
         SOP            &22,634             &56.64+-0.09\\
         ours           &3,298              &72.23\\
         \bottomrule
    \end{tabular}
\end{table}
\subsection{Error Analysis on Image Classification}
\label{subsec:appendix_error_analysis_classification}
\begin{figure}[!h]
    \centering
    \begin{subfigure}{0.3\textwidth}
         \centering
         \includegraphics[height=20mm, keepaspectratio]{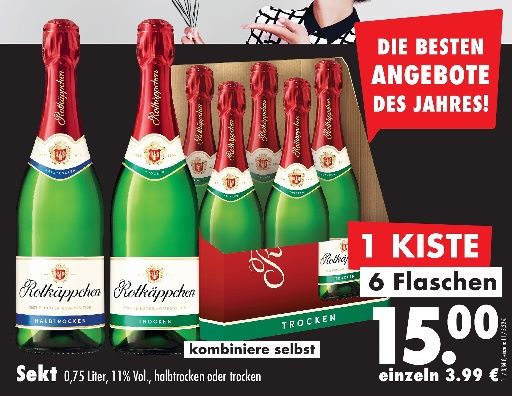}
         \includegraphics[height=20mm, keepaspectratio]{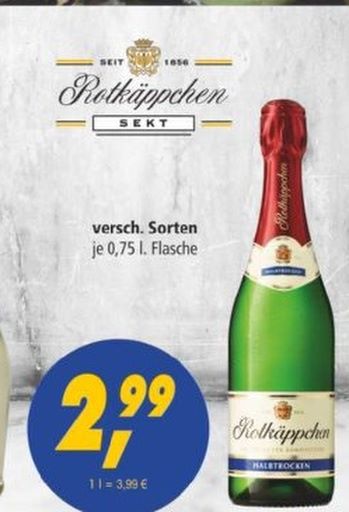}
    \end{subfigure}
    \begin{subfigure}{0.3\textwidth}
        \centering
        \includegraphics[height=18mm, keepaspectratio]{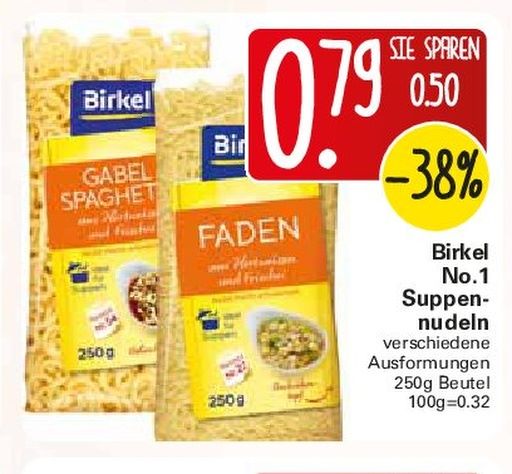}
        \includegraphics[height=18mm, keepaspectratio]{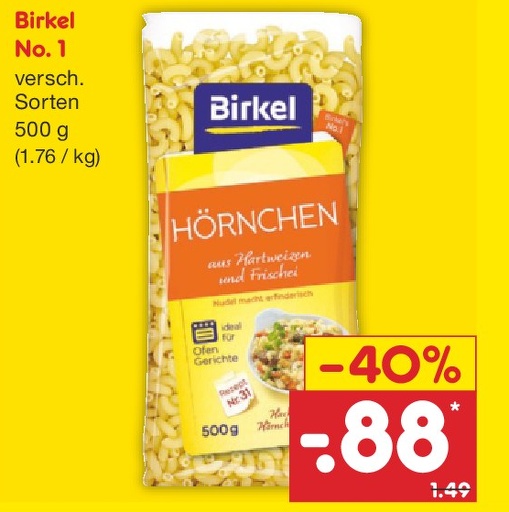}
    \end{subfigure}
    \begin{subfigure}{0.3\textwidth}
        \centering
        \includegraphics[width=\textwidth, height=20mm, keepaspectratio]{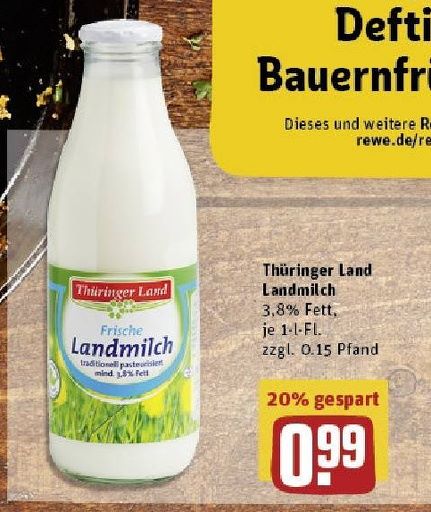}
        \includegraphics[width=\textwidth, height=20mm, keepaspectratio]{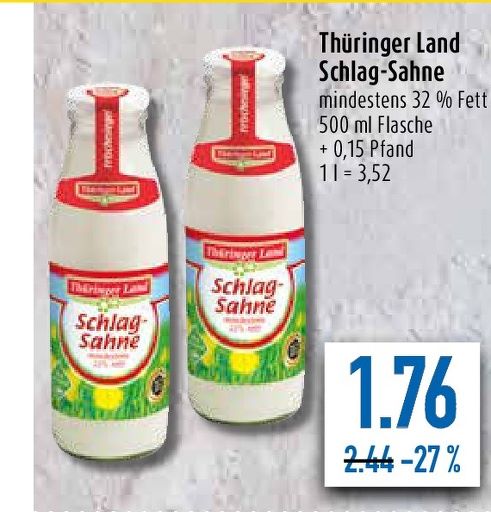}
    \end{subfigure}
    \caption{Example image pairs of failure cases: The images of the entities on the left side are misclassified to the entities illustrated by the images on the right side.
    The images are presented in full resolution from Figure\,\ref{fig:appendix_img_cls_convnext_false_classified_topleft} to Figure\,\ref{fig:appendix_img_cls_convnext_false_classified_bottomright}.
    }
    \label{fig:img_cls_convnext_false_classified}
\end{figure}

Figure\,\ref{fig:img_cls_convnext_false_classified} shows three typical examples of misclassification. The left image of the three subfigures are example images from an entity that are misclassified to the next to it illustrated entity. The left image of the left subfigure indicates the advertisement for a box of six sparkling wine bottles. This entity is misclassified to the entity that only represents only one bottle. Not only the number of products but also the packaging size of a product defines the difference between entities. The centered subfigure illustrated such case. The entity representing the packaging size of 250g is often misclassified to the entity with the packaging size of 500g.
Furthermore, the design of the packaging of products leads to misclassification. The left image of the right subfigure promotes 1l of milk. However, the product is misclassified to the product 1l of cream. Often, there are also a strong difference between the number of the training images of both entities.
For deeper, manual error analysis, we created a much smaller long-tail dataset which consists of 137,416 images in total, where the train split contains 115,889 images, and the test split contains 21,527, respectively. 
Table\,\ref{tab:small_dataset_manual_error_analysis} shows the main reasons for false predictions based on a manual evaluation.
\begin{table}[htb]
    \centering
    \caption{Manual evaluation [in \%] of false classified samples in a subset 21k images.
    }
    \label{tab:small_dataset_manual_error_analysis}
    \begin{tabularx}{\columnwidth}{cccccccc}
        \\
        \toprule
        \begin{tabular}{@{}c@{}}
        different\\size or\\quantity
        \end{tabular} 
        
        &\begin{tabular}{@{}c@{}}
        different\\training\\ and test\\data
        \end{tabular}
        
        &unknown 
        
        &\begin{tabular}{@{}c@{}}
        similar\\design
        \end{tabular}

        &\begin{tabular}{@{}c@{}}
        special\\illustration
        \end{tabular}

        &\begin{tabular}{@{}c@{}}
        focus on\\product
        \end{tabular}

        &\begin{tabular}{@{}c@{}}
        poor image\\quality
        \end{tabular}

        &\begin{tabular}{@{}c@{}}
        false\\GT label
        \end{tabular}
        \\
        \midrule
        $39.0$ &$14.4$ &$10.6$ &$9.7$ &$8.9$&$7.2$ &$6.8$ &$3.4$\\
         \bottomrule
    \end{tabularx}
\end{table}

\newpage
\subsection{Error Analysis on Image Retrieval}
\label{subsec:appendix_error_analysis_retrieval}
\begin{figure}[!h]
     \centering
     \includegraphics[width=0.8\textwidth]{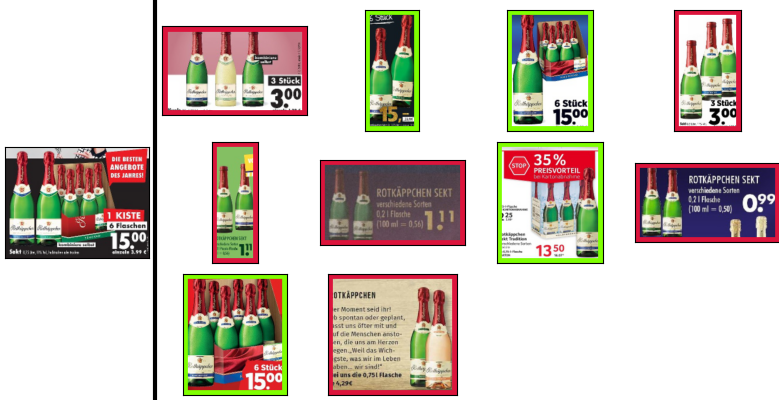}
     \caption{
     Illustration of retrieval sample. The green-framed images belong to the same entity as the query image. The red-framed images belong to another entity. This image query illustrates the lack of the investigated image retrieval model. The most predictions and especially the "nearest" image in the embedding space to the query image belong to another entity.
     The image is presented in full resolution in Figure\,\ref{fig:appendix_error_analysis_image_retrieval}.
     }
     \label{fig:error_analysis_image_retrieval}
\end{figure}

\begin{table}[!h]
    \centering
    \caption{Overview of different datasets described in Section\,\ref{sec:related_work} and in Section\,\ref{sec:appendix_add_realted_work} with the number of images and number of classes per dataset. Furthermore, we indicate whether the dataset is suitable for the FC or EM tasks. Also, we define the image category for the data, if possible. Images recorded in controlled environment are grouped into the staged "studio" image category (studio). Other images belong to the "in the wild" image category (wild). (*) Number of advocated images in \cite{wilke2021towards}, but the dataset could contain more images.}
    \begin{tabularx}{0.855\textwidth}{lcccccc}
    \\ 
    \toprule
         {} & {} & {} & {} & {} &\multicolumn{2}{c}{image category}\\ 
         & \# images & \# classes & FC & EM &studio &wild\\
         \midrule
         \textbf{Online Shopping Products with Images}& & & & &\\
         \textit{Amazon Review Data} (2018) \cite{ni2019justifying}
         &-         &29           &\cmark &\xmark &\cmark &\xmark\\
         {\it Retail Product Categorisation Dataset} \cite{elayanithottathil2021retail}
         &48,000    &21     &\cmark &\xmark &\cmark &\xmark\\
         {\it Atlas} \cite{umaashankar2019atlas}
         &186,150   &52     &\cmark &\xmark &\cmark &\xmark\\
         {\it iMaterialist} competition at FGVC4 \cite{iMaterialistFGVC2017}
         &50,461    &381    &\cmark &\xmark &\cmark &\xmark\\
         {\it iMaterialist} competition at FGVC5 \cite{iMaterialistFGVC2018}
         &214,028   &128    &\cmark &\xmark &\cmark &\xmark\\
         \textbf{Product Images in the Wild}& & & & &\\
         {\it Grozi-120} \cite{merler2007recognizing}
         &5,649     &120    &\cmark &\xmark &\cmark &\cmark\\
         {\it Grozi-3.2k} \cite{george2014recognizing}
         &9,030     &80     &\cmark &\xmark &\cmark &\cmark\\
         {\it Products-10k} \cite{bai2020products}
         &150,000   &10,000 &\cmark &\xmark &\cmark &\cmark\\
         {\it AliProducts Challenge} \cite{Le_2020_ECCV}
         &3,000,000 &50,000 &\cmark &\xmark &\xmark &\cmark\\
         {\it SHORT} \cite{rivera2014small}
         &134,524   &30     &\cmark &\xmark &\cmark &\cmark\\
         \textbf{Shelf Images}& & & & &\\
         Varol and Salih \cite{varol2015toward}
         &5,000     &10     &\cmark &\xmark &\xmark &\cmark\\
         {\it Freiburg Groceries Dataset} \cite{jund2016freiburg}
         &5,021     &25     &\cmark &\xmark &\xmark &\cmark\\
         {\it Retail-121} \cite{karlinsky2017fine}
         &688   &121    &\cmark &\xmark     &\xmark &\cmark\\
         {\it GameStop} \cite{karlinsky2017fine}
         &1,039 &3,700   &\cmark &\xmark &\xmark &\cmark\\
         {\it RP2K} \cite{peng2020rp2k}
         &2,388      &394,696   &\cmark &\xmark &\xmark &\cmark\\
         {\it SKU-110} \cite{goldman2019precise}
         &11,762    &110,712    &\cmark &\xmark &\xmark &\cmark\\
         {\it Locount} \cite{Cai2020Locount}
         &50,394    &140    &\cmark &\xmark &\xmark &\cmark\\
         \textbf{Product Checkout Images}& & & & &\\
         {\it RPC} \cite{wei2022rpc}
         &83,739    &200    &\cmark &\xmark &\cmark &\xmark\\
         {\it ARC-100} \cite{bukhari2021arc}
         &31,000    &100    &\cmark &\xmark &\cmark &\xmark\\
         \textbf{Our Retail-786k dataset}
         &786,179 &3,298 &\xmark &\cmark &\cmark &\xmark\\
         \bottomrule
    \end{tabularx}
    \label{tab:appendix_relatedWork_01}
\end{table}

\begin{table}[!h]
    \centering
    \caption{Continuation of the overview of different datasets described in Section\,\ref{sec:related_work} and in Section\,\ref{sec:appendix_add_realted_work} with the number of images and number of classes per dataset. Furthermore, we indicate whether the dataset is suitable for the FC or EM tasks. Also, we define the image category for the data, if possible. Images recorded in controlled environment are grouped into the staged "studio" image category (studio). Other images belong to the "in the wild" image category (wild). (*) Number of advocated images in \cite{wilke2021towards}, but the dataset could contain more images.}
    \begin{tabularx}{0.865\textwidth}{lcccccc}
    \\ 
    \toprule
         {} & {} & {} & {} & {} &\multicolumn{2}{c}{image category}\\ 
         & \# images & \# classes & FC & EM &studio &wild\\
         \midrule
         \textbf{Products in Vending Machines}& & & & &\\
         {\it TGFS} \cite{hao2019take}
         &38,000    &24     &\cmark &\xmark &\xmark &\cmark\\
         Fuchs et al.\,\cite{fuchs2019towards}
         &300       &90     &\cmark &\xmark &\xmark &\cmark\\
         {\it HoloSelecta} \cite{fuchs2020holoselecta}
         &295       &109    &\cmark &\xmark &\xmark &\cmark\\
         \textbf{Data Bases with Food Images}& & & & &\\
         Rocha  et al.\,\cite{rocha2010automatic}
         &2,633     &15     &\cmark &\xmark &\cmark &\xmark\\
         {\it Food-5K} \cite{singla2016food} 
         &5,000     &2      &\cmark &\xmark &\xmark &\cmark\\
         {\it Food-11} \cite{singla2016food}
         &16,643    &11      &\cmark &\xmark &\xmark &\cmark\\
         \textbf{Miscellaneous}& & & & &\\
         {\it MVTec D2S} \cite{follmann2018mvtec}
         &21,000    &60     &\cmark &\xmark &\cmark &\xmark\\
         {\it SOIL-47} \cite{burianekSOIL47}
         &987       &47     &\cmark &\xmark &\cmark &\xmark\\        
         \textbf{Product Entity Matching}& & & & &\\
         \textit{Benchmark datasets for entity resolution}\cite{kopcke2010evaluation}
         &0         &\textbf{-}      &\xmark &\cmark &\textbf{-} &\textbf{-}\\
         {\it The Magellan Data Repository} \cite{magellandata}
         &0         &\textbf{-}      &\xmark &\cmark &\textbf{-} &\textbf{-}\\
         \textit{Flickr30K Entities} \cite{plummer2015flickr30k}
         &31,783    &\textbf{-}     &\xmark &\cmark &\textbf{-} &\textbf{-}\\
         Wilke and Rahm \cite{wilke2021towards}
         &1,763* &1,763 &\xmark &\cmark &\cmark &\xmark \\
         {\it Stanford Online Products} \cite{oh2016deep}
         &120,053   &22,634 &\xmark &\cmark &\cmark &\xmark\\
         \textbf{Our Retail-786k dataset}
         &786,179 &3,298 &\xmark &\cmark &\cmark &\xmark\\
         \bottomrule
    \end{tabularx}
    \label{tab:appendix_relatedWork_02}
\end{table}
\newpage
\section{Additional Images}

\begin{figure}[!ht]
     \centering
     \begin{subfigure}[b]{0.3\textwidth}
         \centering
         \includegraphics[width=\textwidth]{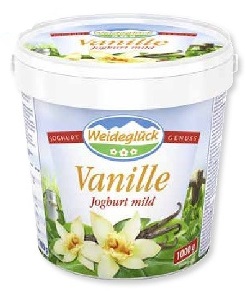}
         \caption{04028900007284}
     \end{subfigure}
     \hspace{2mm}
     \begin{subfigure}[b]{0.3\textwidth}
         \centering
         \includegraphics[width=\textwidth]{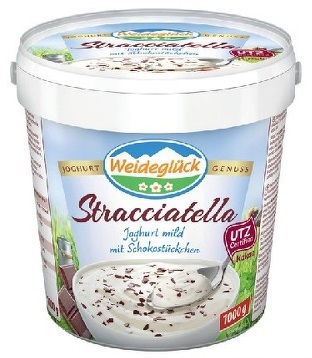}
         \caption{04028900007413}
     \end{subfigure}
     \begin{subfigure}[b]{0.3\textwidth}
         \centering
         \includegraphics[width=\textwidth]{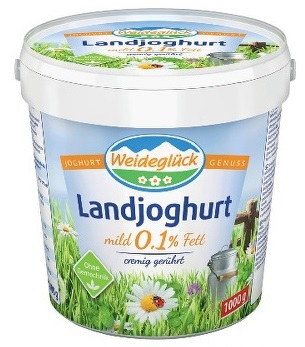}
         \caption{04028900007321}
     \end{subfigure}
        \caption{Illustration of three different products, each product has its own GTIN described in the caption. They differ from each other in the flavor but might be used independently by different retailers to promote discounts for all three of them. Hence, they form an entity in a price comparison task. 
        For higher resolution versions, refer to Figure\,\ref{fig:appendix_product_vanille}, \ref{fig:appendix_product_stracciatella}, and \ref{fig:appendix_product_landjoghurt}.
        }
        \label{fig:product_examples}
\end{figure}
\begin{figure}[!ht]
     \centering
     \begin{subfigure}[b]{0.3\textwidth}
         \centering
         \includegraphics[width=\textwidth]{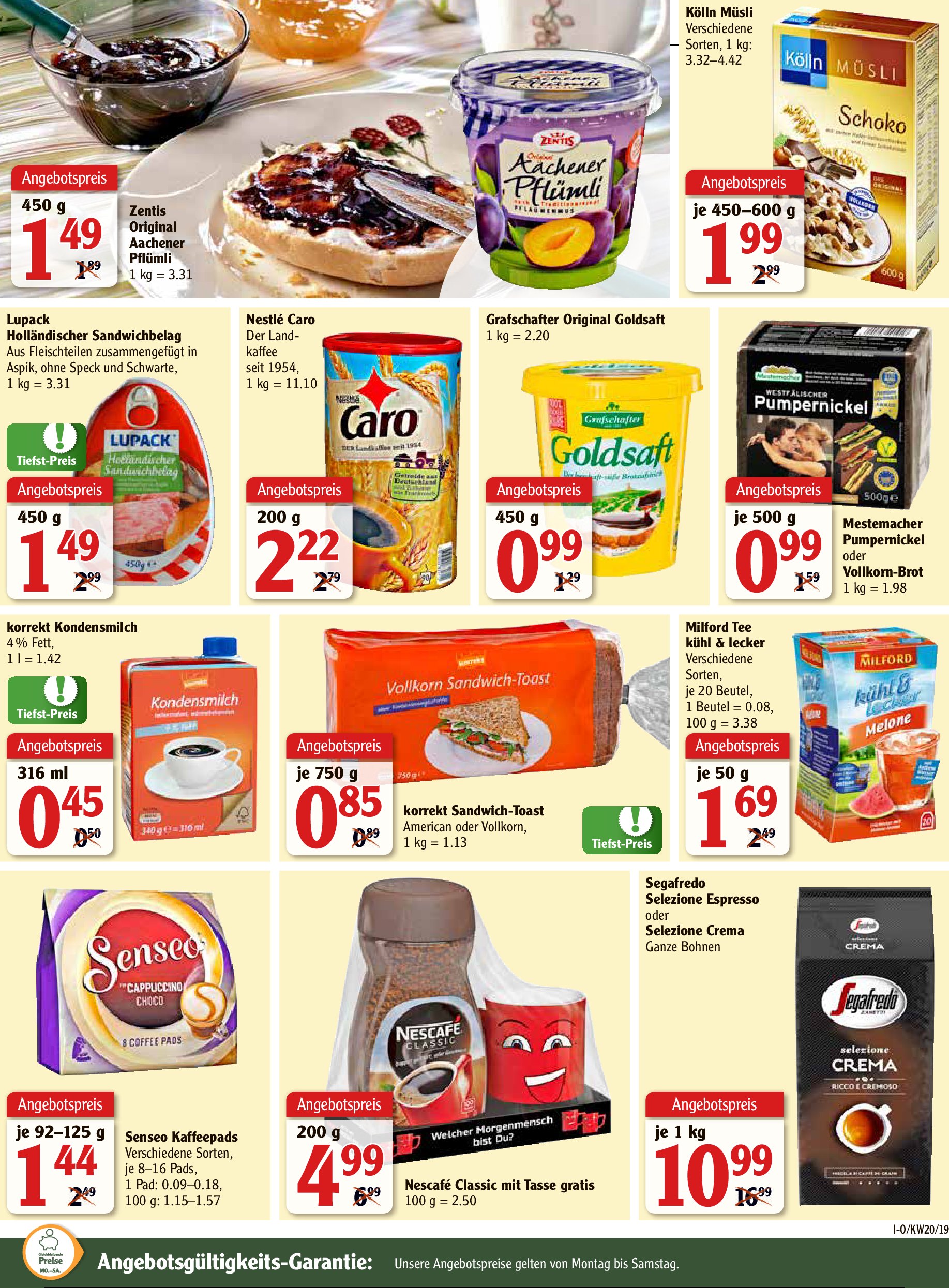}
     \end{subfigure}
     \hfill
     \begin{subfigure}[b]{0.27\textwidth}
         \centering
         \includegraphics[width=\textwidth]{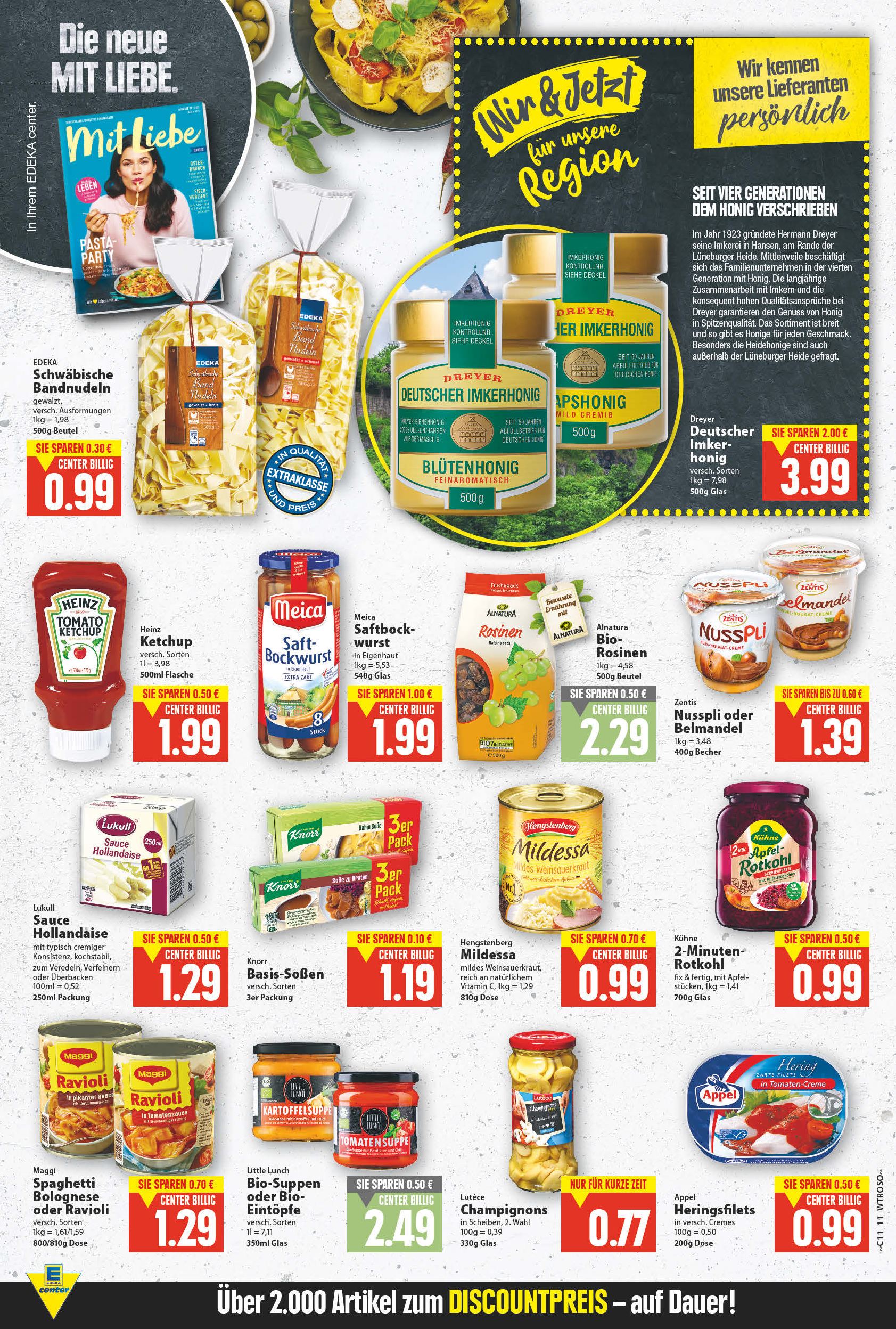}
     \end{subfigure}
     \hfill
     \begin{subfigure}[b]{0.28\textwidth}
         \centering
         \includegraphics[width=\textwidth]{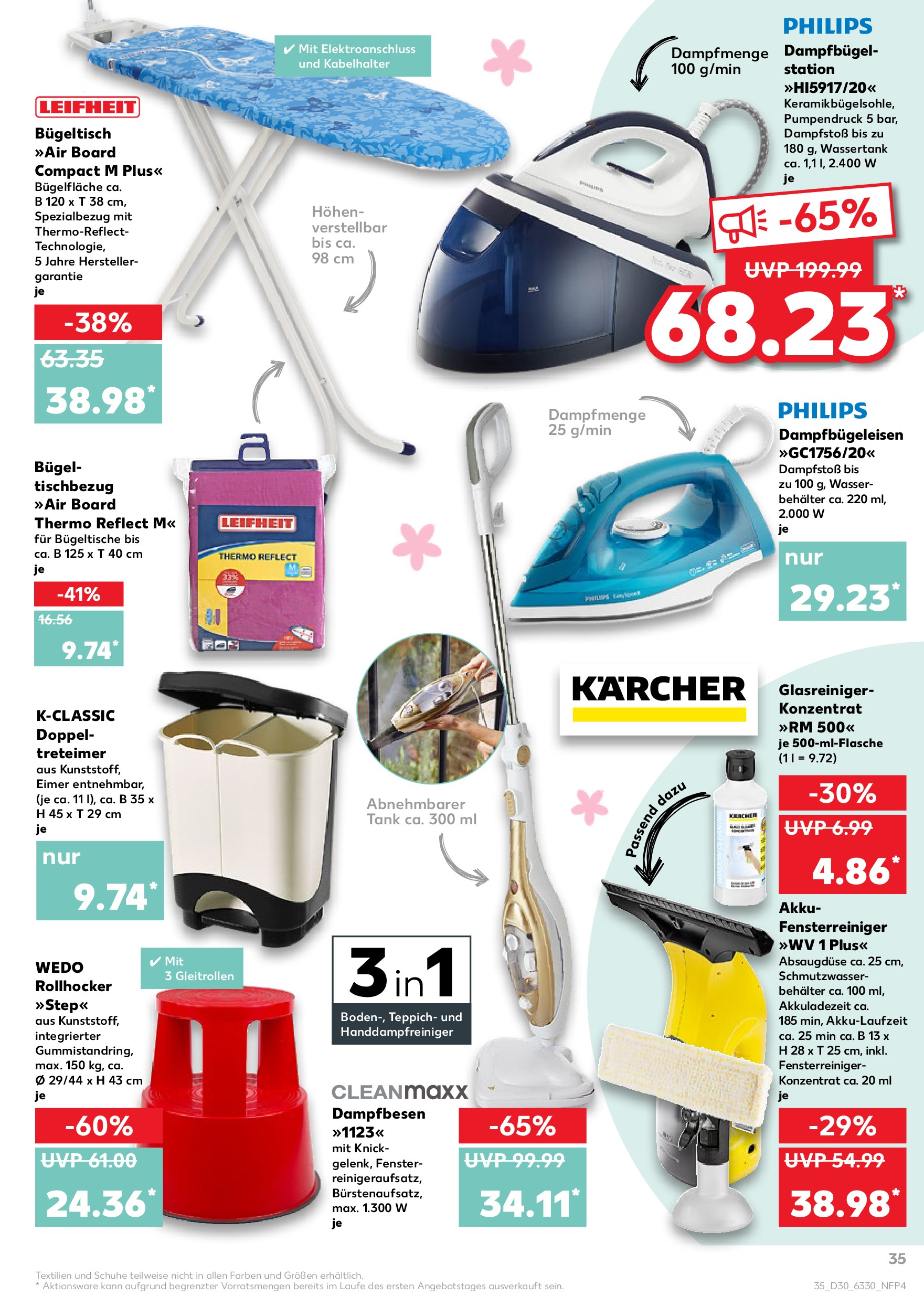}
     \end{subfigure}
        \caption{Examples of leaflet pages from different European retailers which are the basis of the proposed dataset.
        For higher resolution versions, refer to  Figure\,\ref{fig:appendix_leaflet_page_Globus}, \ref{fig:appendix_leaflet_page_Ecenter}, and \ref{fig:appendix_leaflet_page_Kaufland}.
        }
        \label{fig:leaflet_examples}
\end{figure}
\begin{figure}[!h]
     \centering
     \includegraphics[width=\textwidth]  {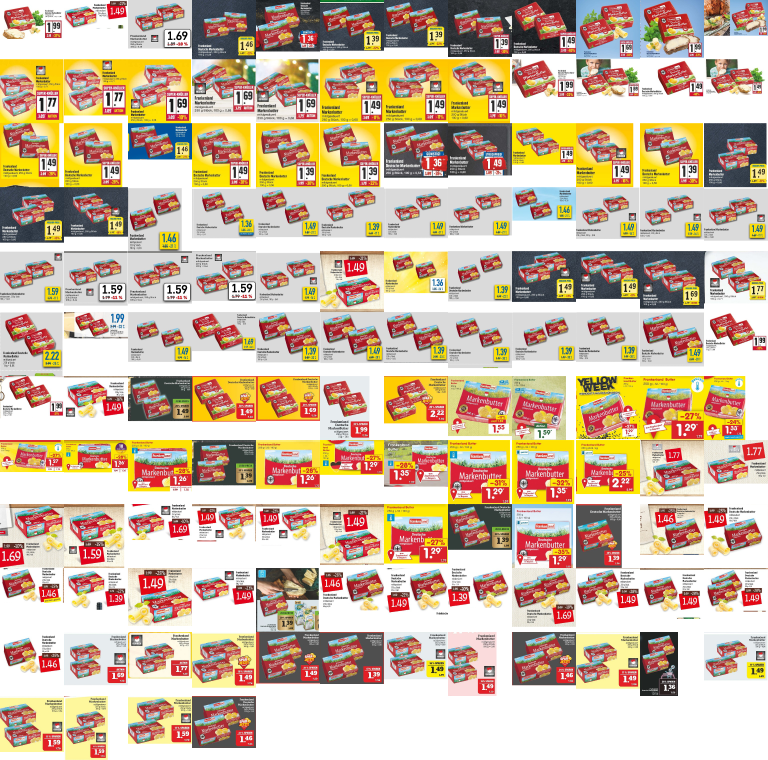}
    \caption{The training set of the entity ID 28 is depicted. Figure\,\ref{fig:appendix_test_set} shows the corresponding test set.}
    \label{fig:appendix_train_set}
\end{figure}
\begin{figure}[!h]
     \centering
     \includegraphics[width=\textwidth]{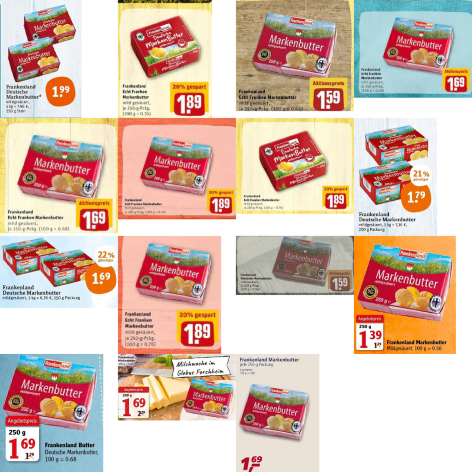}
    \caption{The test set of the entity ID 28 is depicted. Figure\,\ref{fig:appendix_train_set} shows the corresponding training set.}
    \label{fig:appendix_test_set}
\end{figure}
\begin{figure}
    \centering
    \includegraphics[width=\textwidth]{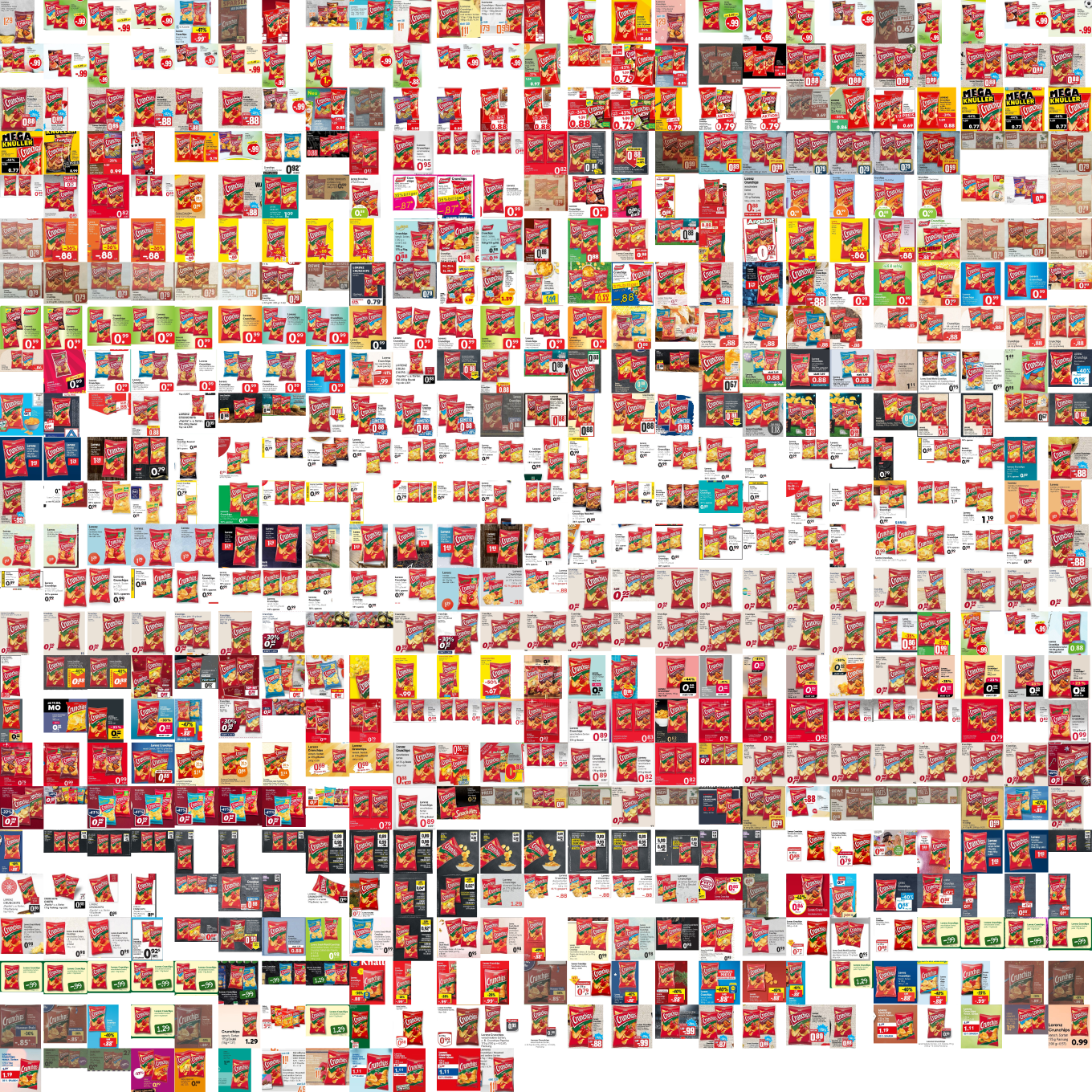}
    \caption{The training set of the entity ID 18 is depicted.}
    \label{fig:appendix_similarity_Chips_CrunchChips}
\end{figure}
\begin{figure}
    \centering
    \includegraphics[width=\textwidth]{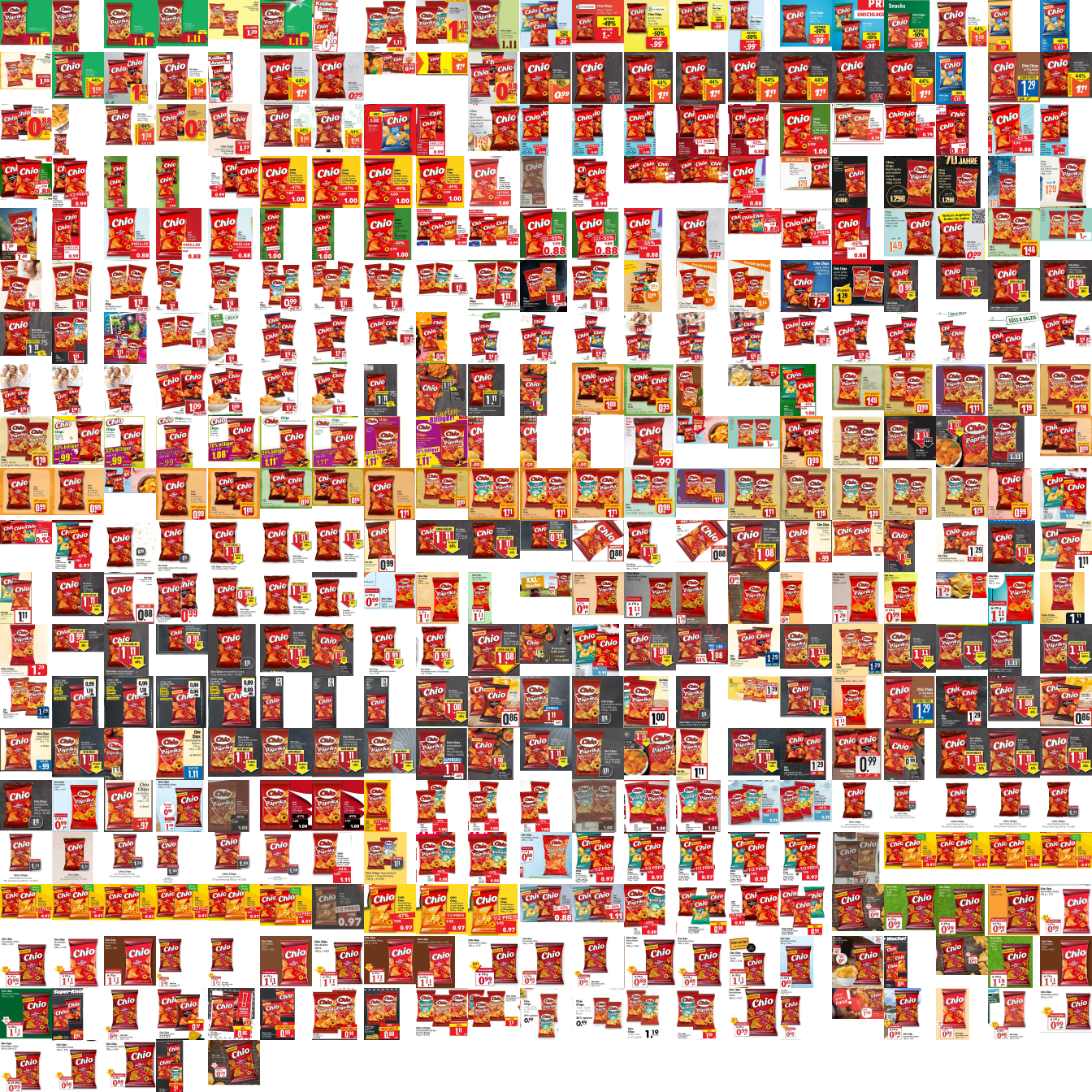}
    \caption{The training set of the entity ID 51 is depicted.}
    \label{fig:appendix_similarity_Chips_Chio}
\end{figure}
\begin{sidewaysfigure}
    \centering
    \includegraphics[width=\textwidth]{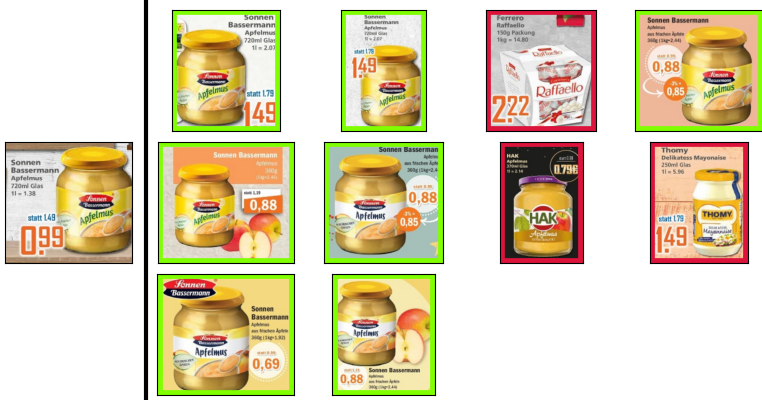}
    \caption{A further example for image retrieval with the product "apple sauce". This entity differs in the brand as the result of the right figure of Figure\,\ref{fig:image_retrieval_Chips_Apfelmus}.}
    \label{fig:appendix_image_retrieval_query_01}
\end{sidewaysfigure}
\begin{sidewaysfigure}
    \centering
    \includegraphics[width=\textwidth]{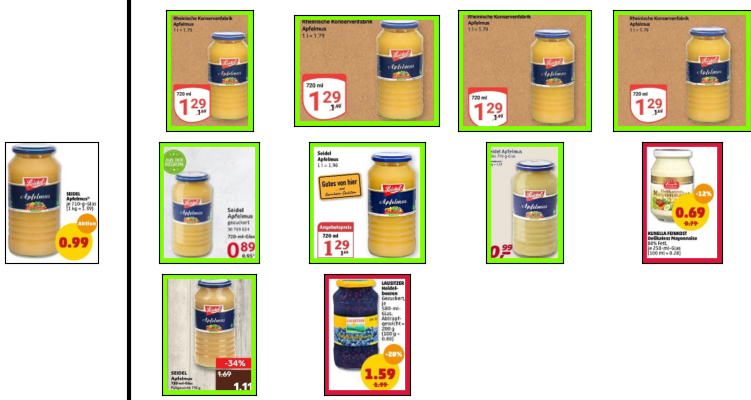}
    \caption{A further example for image retrieval with the product "apple sauce". This entity differs in the packaging as the result of the right figure of Figure\,\ref{fig:image_retrieval_Chips_Apfelmus} and Figure\,\ref{fig:appendix_image_retrieval_query_01}.}
    \label{fig:appendix_image_retrieval_query_02}
\end{sidewaysfigure}
\begin{sidewaysfigure}
    \centering
    \includegraphics[width=\textwidth]{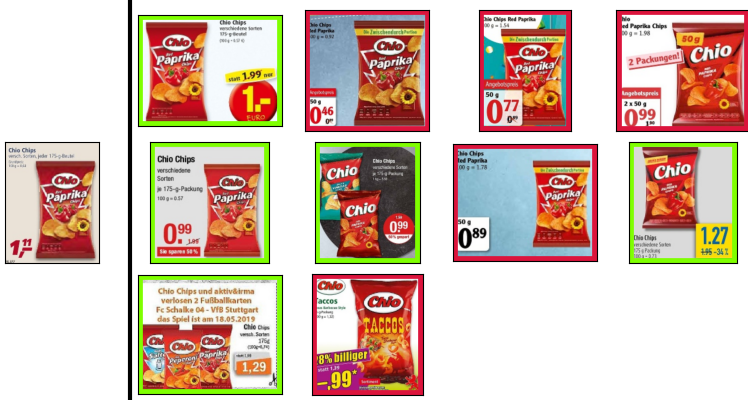}
    \caption{Illustration of an image retrieval with the product "potato chips". The query image is from the test set of the entity ID 51. The training set is shown in Figure\,\ref{fig:appendix_similarity_Chips_Chio}.}
    \label{fig:appendix_image_retrieval_query_03}
\end{sidewaysfigure}

\noindent Example images of a training and a test set of the entity ID 28 are shown in Figure\,\ref{fig:appendix_train_set} and Figure\,\ref{fig:appendix_test_set}, respectively.\\
\\
A further example for strong similarity between the images of different entities are shown in Figure\,\ref{fig:appendix_similarity_Chips_CrunchChips} and Figure\,\ref{fig:appendix_similarity_Chips_Chio}.\\
\\
Furthermore, more image retrieval results for different query images are generated. Figure\,\ref{fig:appendix_image_retrieval_query_01}, Figure\,\ref{fig:appendix_image_retrieval_query_02}, and Figure\,\ref{fig:appendix_image_retrieval_query_03}, show the results.



\begin{sidewaysfigure}
    \centering
    \def\svgwidth{\columnwidth}
    \input{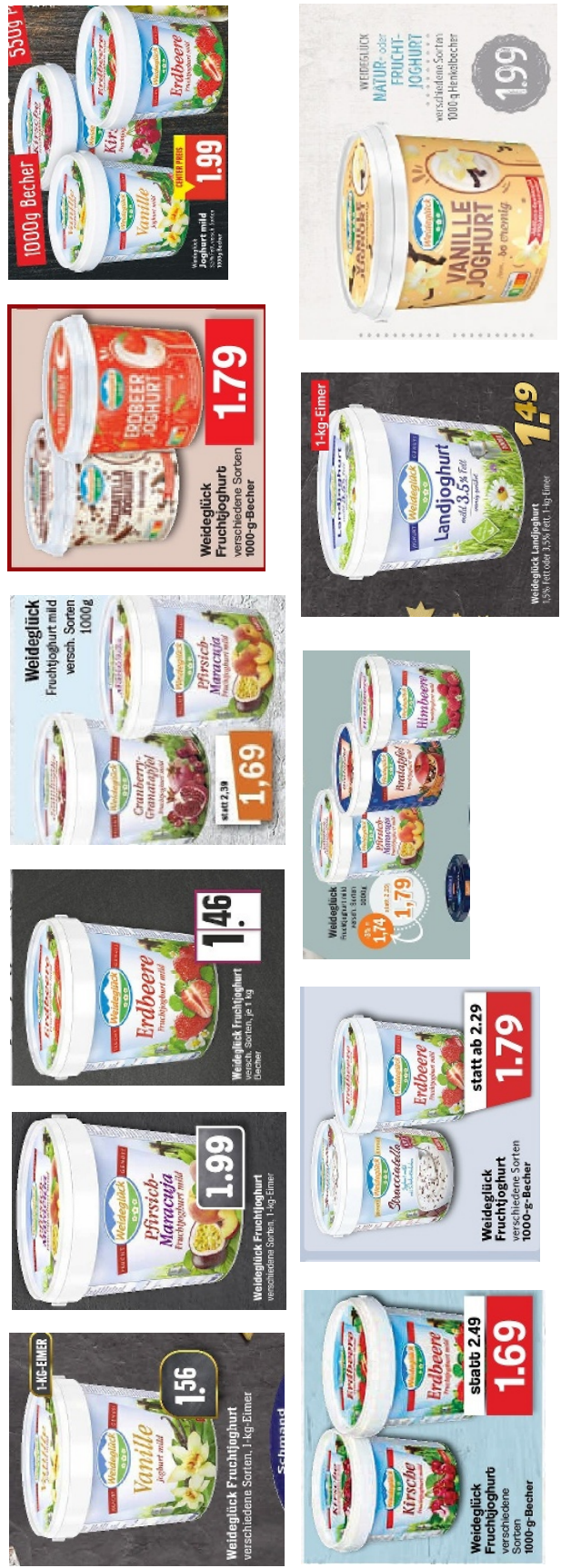}
    \caption{Illustration of the Figure\,\ref{fig:visual_abstract_explain} in full resolution.}
    \label{fig:appendix_visual_abstract_explain}
\end{sidewaysfigure}

\begin{figure}
     \centering
     \includegraphics[width=\textwidth]{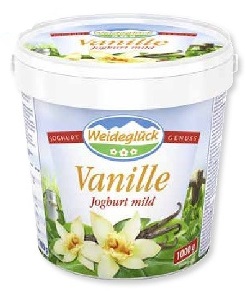}
    \caption{Illustration of the left product image in Figure\,\ref{fig:product_examples} in full resolution.}
    \label{fig:appendix_product_vanille}
\end{figure}

\begin{figure}
     \centering
     \includegraphics[width=\textwidth]{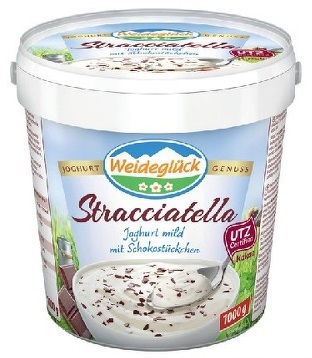}
    \caption{Illustration of the middle product image in Figure\,\ref{fig:product_examples} in full resolution.}
    \label{fig:appendix_product_stracciatella}
\end{figure}

\begin{figure}
     \centering
     \includegraphics[width=\textwidth]{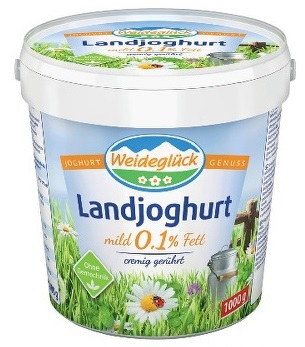}
    \caption{Illustration of the right product image in Figure\,\ref{fig:product_examples} in full resolution.}
    \label{fig:appendix_product_landjoghurt}
\end{figure}

\begin{figure}
     \centering
     \includegraphics[width=\textwidth]{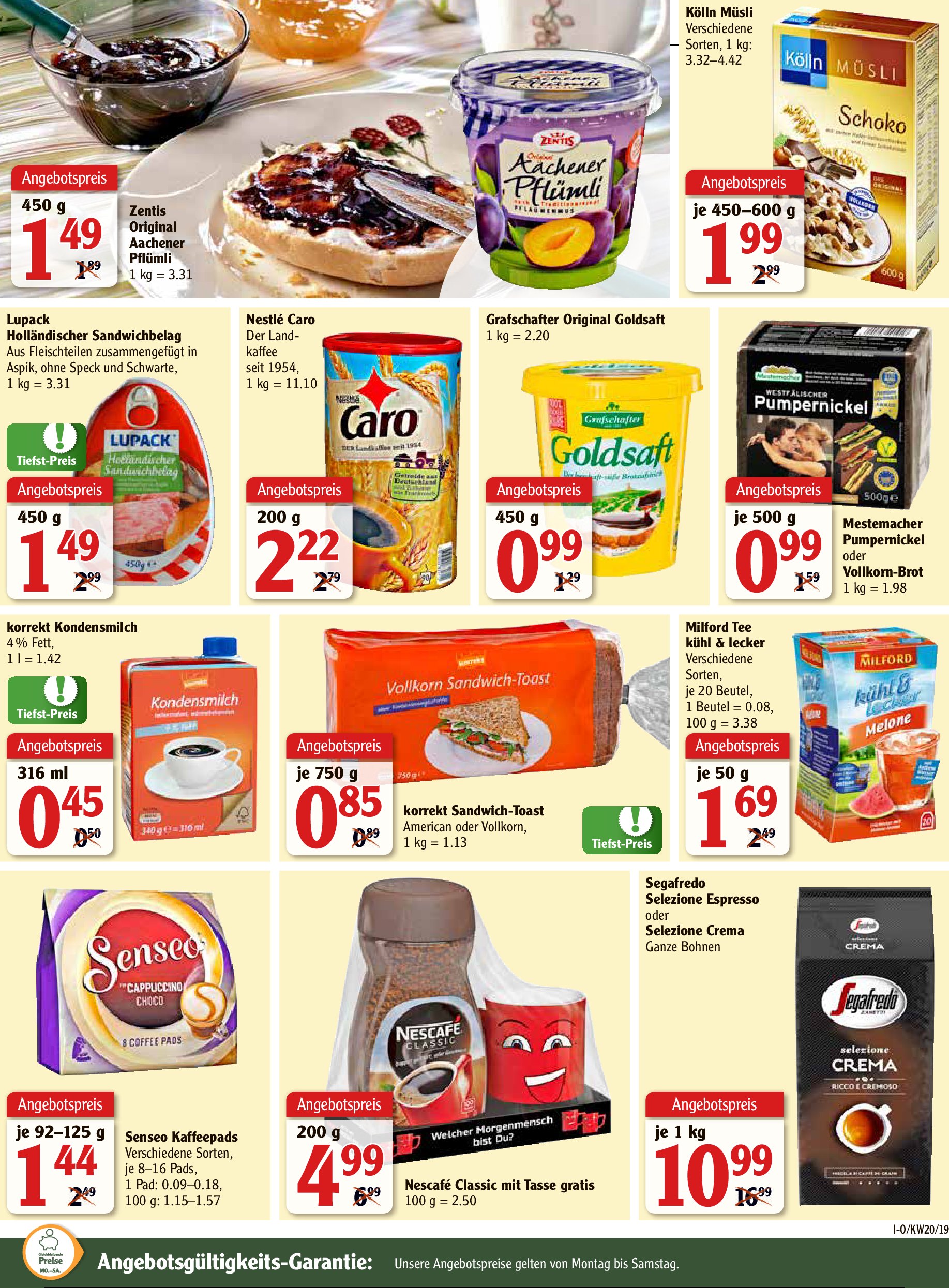}
    \caption{Illustration of the left leaflet page in Figure\,\ref{fig:leaflet_examples} in full resolution.}
    \label{fig:appendix_leaflet_page_Globus}
\end{figure}

\begin{figure}
     \centering
     \includegraphics[width=0.94\textwidth]{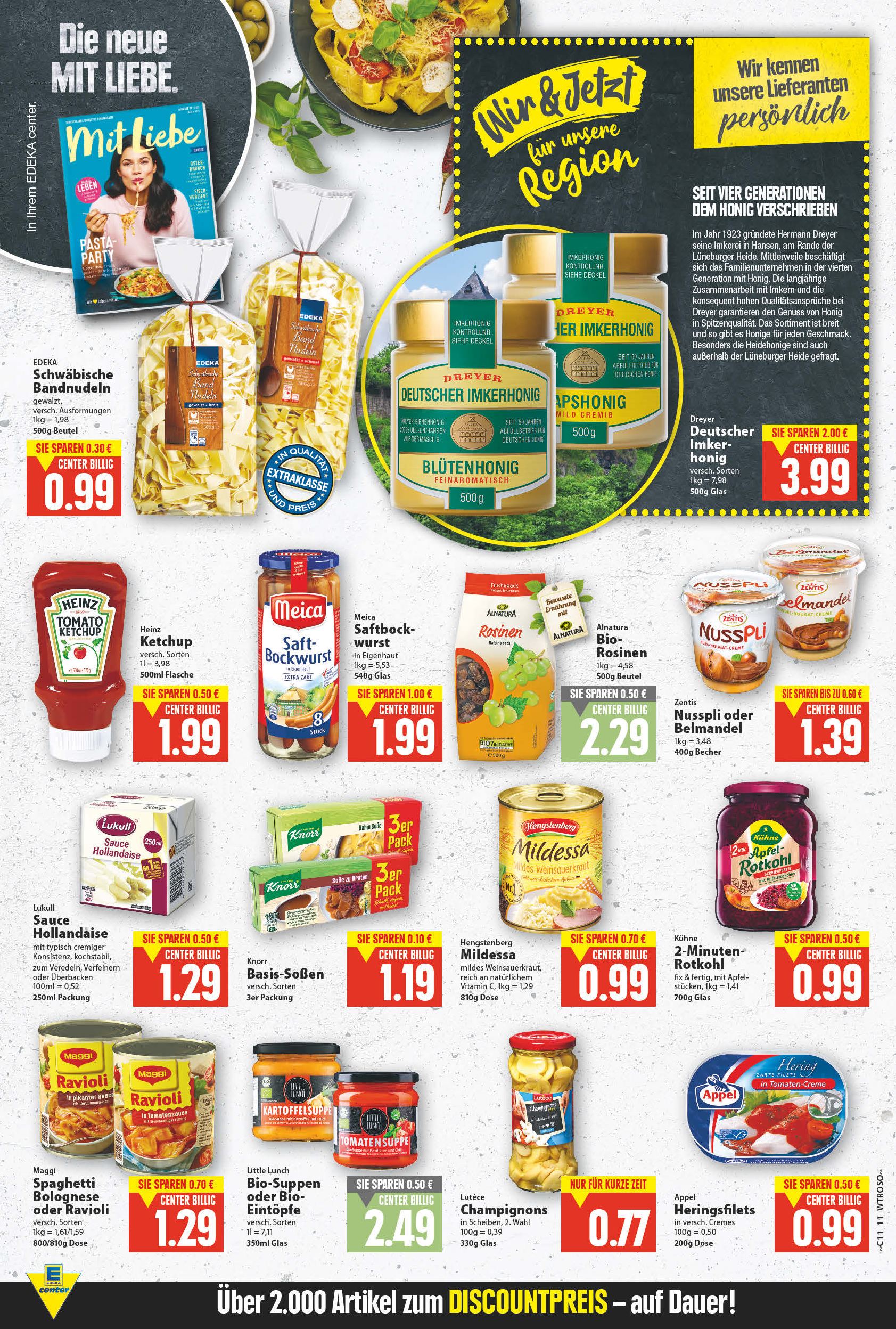}
    \caption{Illustration of the middle leaflet page in Figure\,\ref{fig:leaflet_examples} in full resolution.}
    \label{fig:appendix_leaflet_page_Ecenter}
\end{figure}

\begin{figure}
     \centering
     \includegraphics[width=0.98\textwidth]{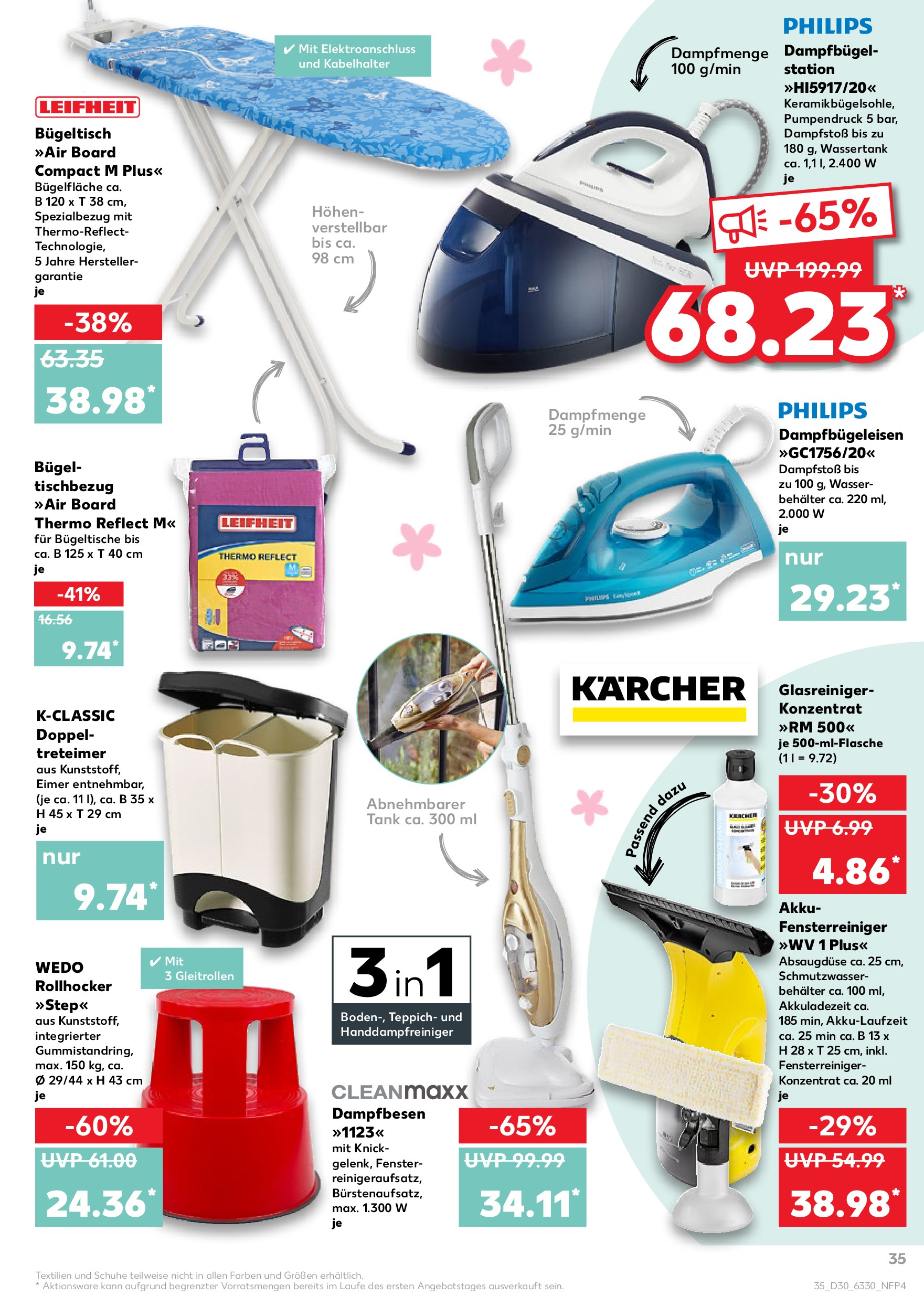}
    \caption{Illustration of the right leaflet page in Figure\,\ref{fig:leaflet_examples} in full resolution.}
    \label{fig:appendix_leaflet_page_Kaufland}
\end{figure}

\begin{figure}
    \centering
    \includegraphics[width=\textwidth]{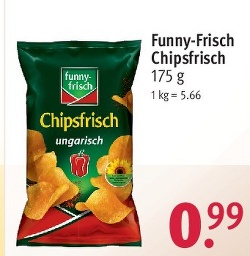}
    \caption{Illustration of the first image of Figure\,\ref{fig:entity_example_chips} in full resolution.}
    \label{fig:appendix_chips_175g}
\end{figure}

\begin{figure}
    \centering
    \includegraphics[width=\textwidth]{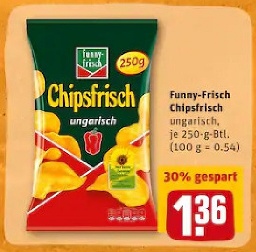}
    \caption{Illustration of the second image of Figure\,\ref{fig:entity_example_chips} in full resolution.}
    \label{fig:appendix_chips_250g}
\end{figure}

\begin{figure}
    \centering
    \includegraphics[width=\textwidth]{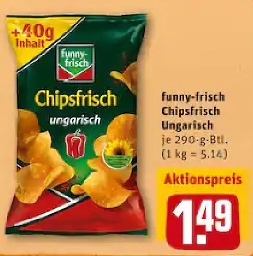}
    \caption{Illustration of the third image of Figure\,\ref{fig:entity_example_chips} in full resolution.}
    \label{fig:appendix_chips_290g}
\end{figure}

\begin{figure}
    \centering
    \includegraphics[width=\textwidth]{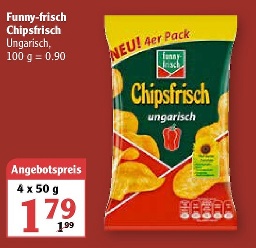}
    \caption{Illustration of the fourth image of Figure\,\ref{fig:entity_example_chips} in full resolution.}
    \label{fig:appendix_chips_4x50g}
\end{figure}

\begin{figure}
    \centering
    \includegraphics[width=\textwidth]{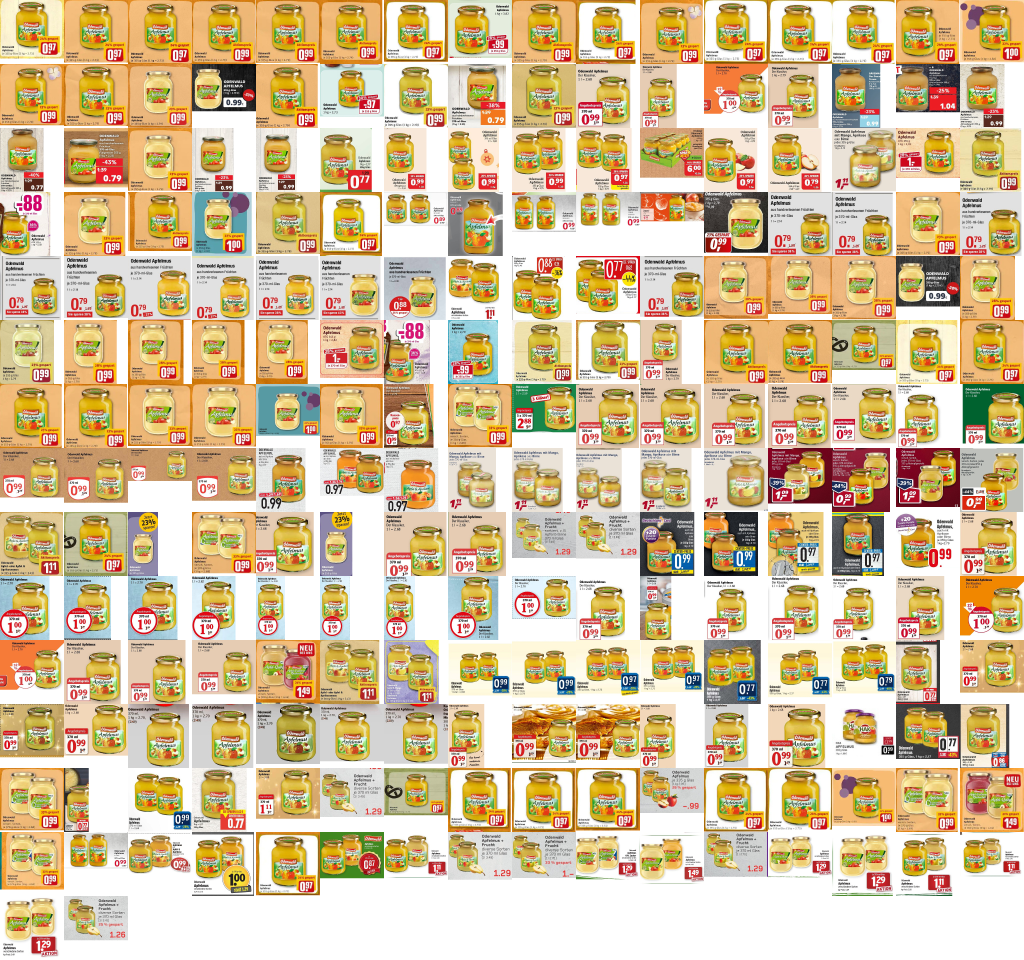}
    \caption{The training set of the entity ID 371 is depicted.}
    \label{fig:appendx_similarity_Apfelmuss_Odenwald}
\end{figure}

\begin{figure}
    \centering
    \includegraphics[width=\textwidth]{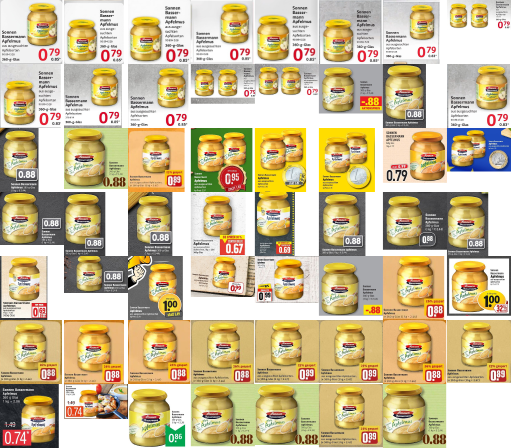}
    \caption{The training set of the entity ID 373 is depicted.}
    \label{fig:appendx_similarity_Apfelmuss_SonnenBassermann}
\end{figure}

\begin{figure}
    \centering
    \includegraphics[width=\textwidth]{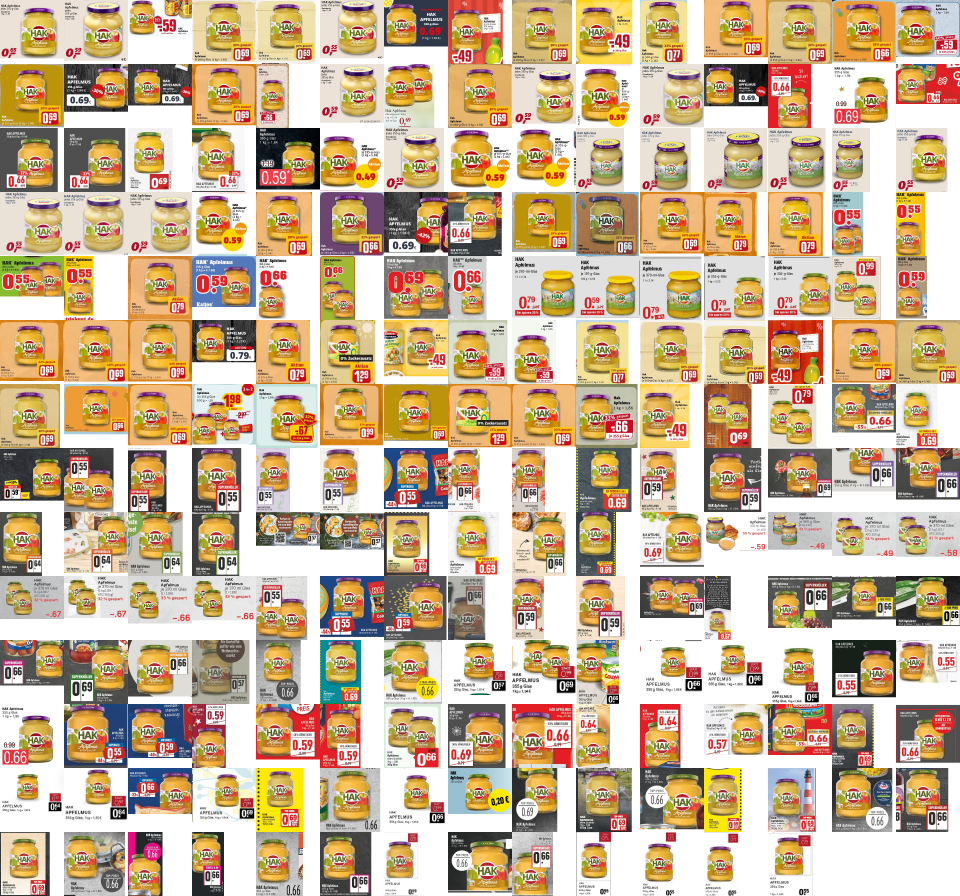}
    \caption{The training set of the entity ID 1208 is depicted.}
    \label{fig:appendx_similarity_Apfelmus_HAK}
\end{figure}

\begin{sidewaysfigure}
    \centering
    \includegraphics[width=\textwidth]{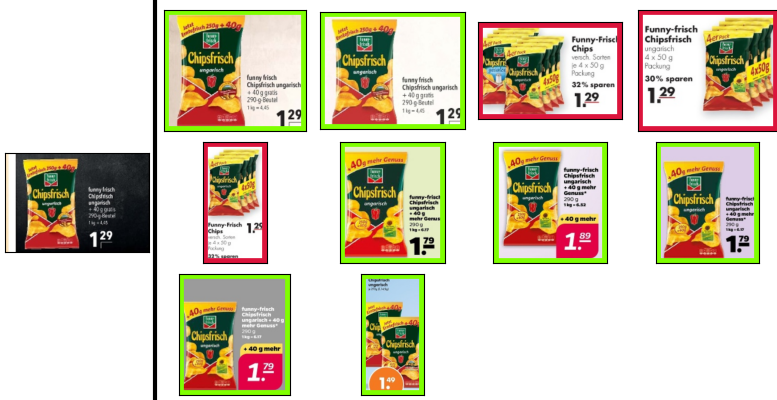}
    \caption{Illustration of the left image of Figure\,\ref{fig:image_retrieval_Chips_Apfelmus} in full resolution.}
    \label{fig:appendix_image_retrieval_Chips}
\end{sidewaysfigure}
\begin{sidewaysfigure}
    \centering
    \includegraphics[width=\textwidth]{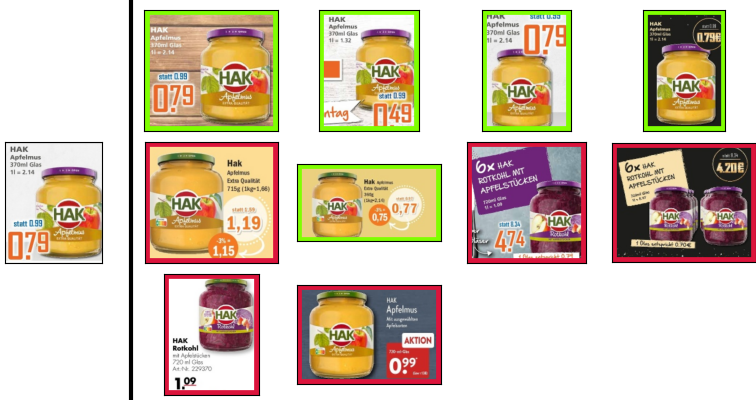}
    \caption{Illustration of the right image of Figure\,\ref{fig:image_retrieval_Chips_Apfelmus} in full resolution.}
    \label{fig:appendix_image_retrieval_Apfelmus} 
\end{sidewaysfigure}




\begin{figure}
    \centering
    \includegraphics[width=\textwidth]{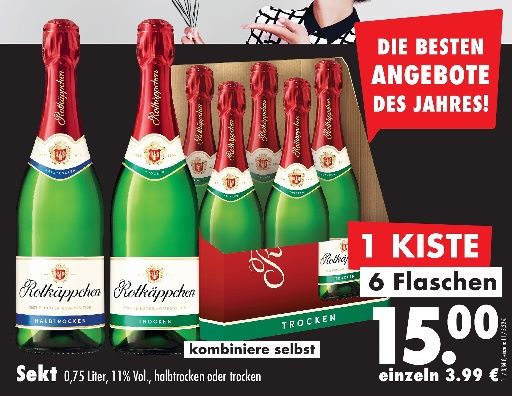}
    \caption{Illustration of the top left image of Figure\,\ref{fig:img_cls_convnext_false_classified} in full resolution.}
    \label{fig:appendix_img_cls_convnext_false_classified_topleft}
\end{figure}

\begin{figure}
    \centering
    \includegraphics[width=\textwidth]{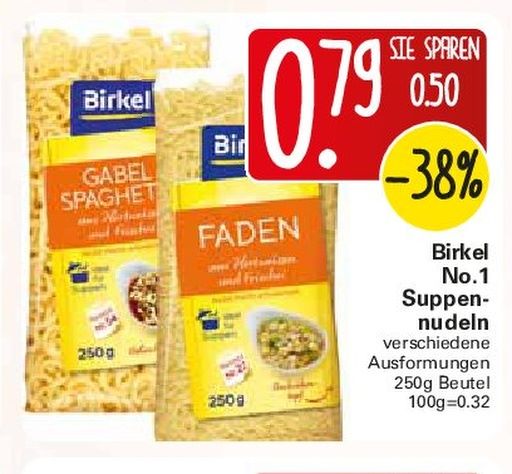}
    \caption{Illustration of the top centered image of Figure\,\ref{fig:img_cls_convnext_false_classified} in full resolution.}
    \label{fig:appendix_img_cls_convnext_false_classified_topcentered}
\end{figure}

\begin{figure}
    \centering
    \includegraphics[width=\textwidth]{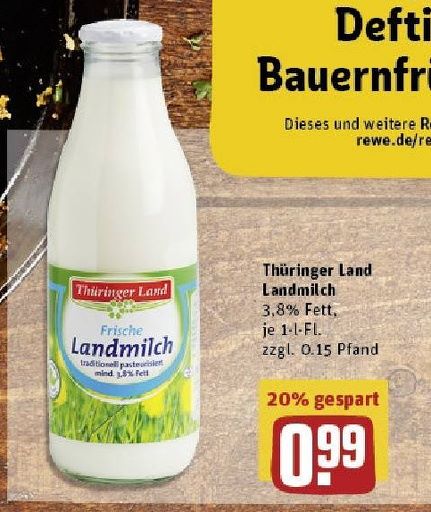}
    \caption{Illustration of the top right image of Figure\,\ref{fig:img_cls_convnext_false_classified} in full resolution.}
    \label{fig:appendix_img_cls_convnext_false_classified_topright}
\end{figure}

\begin{figure}
    \centering
    \includegraphics[width=0.95\textwidth]{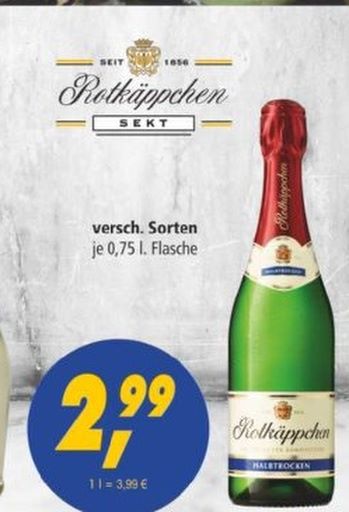}
    \caption{Illustration of the bottom left image of Figure\,\ref{fig:img_cls_convnext_false_classified} in full resolution.}
    \label{fig:appendix_img_cls_convnext_false_classified_bottomleft}
\end{figure}

\begin{figure}
    \centering
    \includegraphics[width=\textwidth]{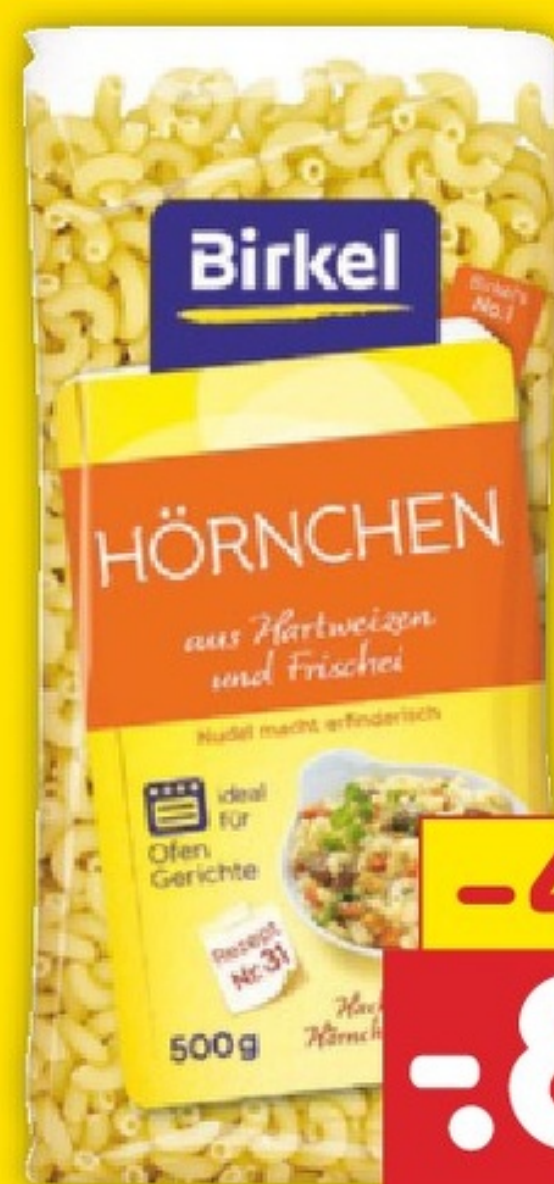}
    \caption{Illustration of the bottom centered image of Figure\,\ref{fig:img_cls_convnext_false_classified} in full resolution.}
    \label{fig:appendix_img_cls_convnext_false_classified_bottom centered}
\end{figure}

\begin{figure}
    \centering
    \includegraphics[width=\textwidth]{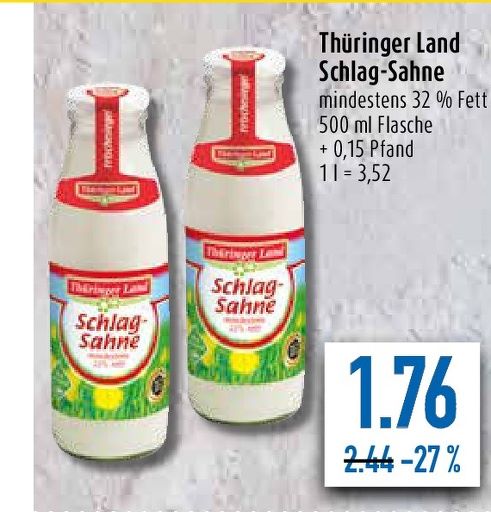}
    \caption{Illustration of the bottom right image of Figure\,\ref{fig:img_cls_convnext_false_classified} in full resolution.}
    \label{fig:appendix_img_cls_convnext_false_classified_bottomright}
\end{figure}

\begin{sidewaysfigure}
    \centering
    \includegraphics[width=\textwidth]{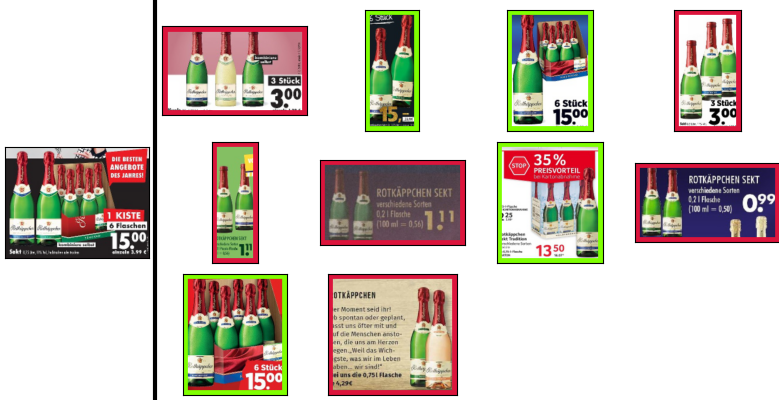}
    \caption{Illustration of an image retrieval with the product "sparkling wine". The query image is from the test set of the entity ID 430.}
    \label{fig:appendix_error_analysis_image_retrieval}
\end{sidewaysfigure}

\end{document}